\documentclass[10pt,journal,compsoc]{IEEEtran}
\ifCLASSOPTIONcompsoc
  \usepackage[nocompress]{cite}
\else
  \usepackage{cite}
\fi
\ifCLASSINFOpdf
\else
\fi
\usepackage{epsfig}
\usepackage{graphicx}
\usepackage{amsmath}
\usepackage{amssymb}
\usepackage{algpseudocode}
\usepackage{algorithm}

\usepackage{xcolor}
\usepackage{verbatim}
\usepackage{multirow}
\usepackage{makecell}
\usepackage{wrapfig}
\usepackage{caption}
\usepackage{array}
\usepackage{bbold}
\usepackage[pagebackref=true,breaklinks=true,colorlinks,bookmarks=false]{hyperref}

\newcommand{\f}{\boldsymbol{f}}

\newcommand{\z}{\boldsymbol{z}}

\newcommand{\h}{\boldsymbol{h}}

\newcommand{\W}{\boldsymbol{W}}

\renewcommand{\v}{\boldsymbol{v}}

\renewcommand{\H}{\boldsymbol{H}}
\DeclareMathOperator{\E}{\mathbb{E}}

\newcommand{\mL}{\mathcal{L}}
\newcommand{\C}{\mathcal{C}}

\newcommand{\U}{\mathcal{U}}
\newcommand{\Q}{\mathrm{Q}}

\newcommand{\argmin}{\operatorname{argmin}}
\newcommand{\argmax}{\operatorname{argmax}}

\newtheorem{remark}{Remark}

\algnewcommand{\LineComment}[1]{\State \(\triangleright\) #1}

\newcommand{\myparagraph}[1]{\vspace{-2pt}\medskip\noindent\textbf{#1}}

\begin{document}
%
\title{Compositional Fine-Grained Low-Shot Learning}

%
%
%
%

\author{Dat~Huynh
        and Ehsan Elhamifar
\IEEEcompsocitemizethanks{\IEEEcompsocthanksitem D.~Huynh and E.~Elhamifar are with the Khoury College of Computer Sciences, Northeastern University, Boston, MA, 02115.\protect\\}
}

\IEEEtitleabstractindextext{%
\begin{abstract}
We develop a novel compositional generative model for zero- and few-shot learning to recognize fine-grained classes with a few or no training samples. Our key observation is that generating holistic features for fine-grained classes fails to capture small attribute differences between classes. Therefore, we propose a feature composition framework that learns to extract attribute features from training samples and combines them to construct fine-grained features for rare and unseen classes. Feature composition allows us to not only selectively compose features of every class from only relevant training samples, but also obtain diversity among composed features via changing samples used for the composition. In addition, instead of building holistic features for classes, we use our attribute features to form dense representations capable of capturing fine-grained attribute details of classes. We propose a training scheme that uses a discriminative model to construct features that are subsequently used to train the model itself. Therefore, we directly train the discriminative model on the composed features without learning a separate generative model.
We conduct experiments on four popular datasets of DeepFashion, AWA2, CUB, and SUN, showing the effectiveness of our method.
\end{abstract}

\begin{IEEEkeywords}
Zero-Shot Learning, Few-Shot Learning, Fine-Grained Recognition, Visual Attention, Compositional Learning
\end{IEEEkeywords}}

\maketitle

\IEEEdisplaynontitleabstractindextext

%
\IEEEpeerreviewmaketitle


%
%
%
%
\section{Introduction}
Fine-grained recognition, which is to classify categories that are visually very similar, is an important yet challenging task with a wide range of applications from fashion industry, e.g., recognition of different types of shoe or cloth \cite{Wang:CVPR18,Liu:CVPR16,Ak:CVPR18}, to face recognition \cite{Parkhi:BMVC15,Wen:ECCV16,Liu:ICCV15} and environmental conservation, e.g., recognizing endangered species of birds or plants \cite{Ding:ICCV19,elhoseiny:CVPR17,Lin:ICCV15,Zheng:ICCV17,Zhao:ICCV19,Zhang:CVPR16}. However, training fine-grained classification systems is challenging, as collecting sufficient training samples from every class requires costly annotations by domain experts to distinguish between similar classes, e.g., `Parakeet Auklet' and `Least Auklet' bird species or `Sweatpant' and `Jeggings', see Figure \ref{fig:att_samples}. Thus, training samples often follow a long-tail distribution \cite{Huynh-mll:CVPR20,Wertheimer:CVPR19}, where many classes have few or no training samples. In this work, we aim to generalize fine-grained recognition to novel classes with a few or no training samples via capturing and transferring fine-grained knowledge from seen to novel classes without overfitting on seen classes.

Although fine-grained classification has achieved remarkable performance by using feature pooling \cite{Kong:CVPR17,Gao:CVPR16,Lin:ICCV15} and discriminative region localization \cite{Zhao:ICCV19,Ding:ICCV19,Ak:CVPR18,Wang:CVPR18,Zhang:CVPR16,Liu:CVPR16} techniques, these works cannot generalize to novel classes, as it requires sufficient training samples from every class, and cannot leverage auxiliary information such as class semantic vectors, which is fundamental for transferring knowledge to novel classes. With the need for costly training data, conventional fine-grained classification methods cannot scale to a large number of classes. However, fine-grained classes can often be described in terms of attributes that are common among classes. Thus, using these semantic descriptions to transfer knowledge among classes can significantly reduce the amount of training samples.

\begin{figure}[t]
\centering
\includegraphics[width=0.99\linewidth]{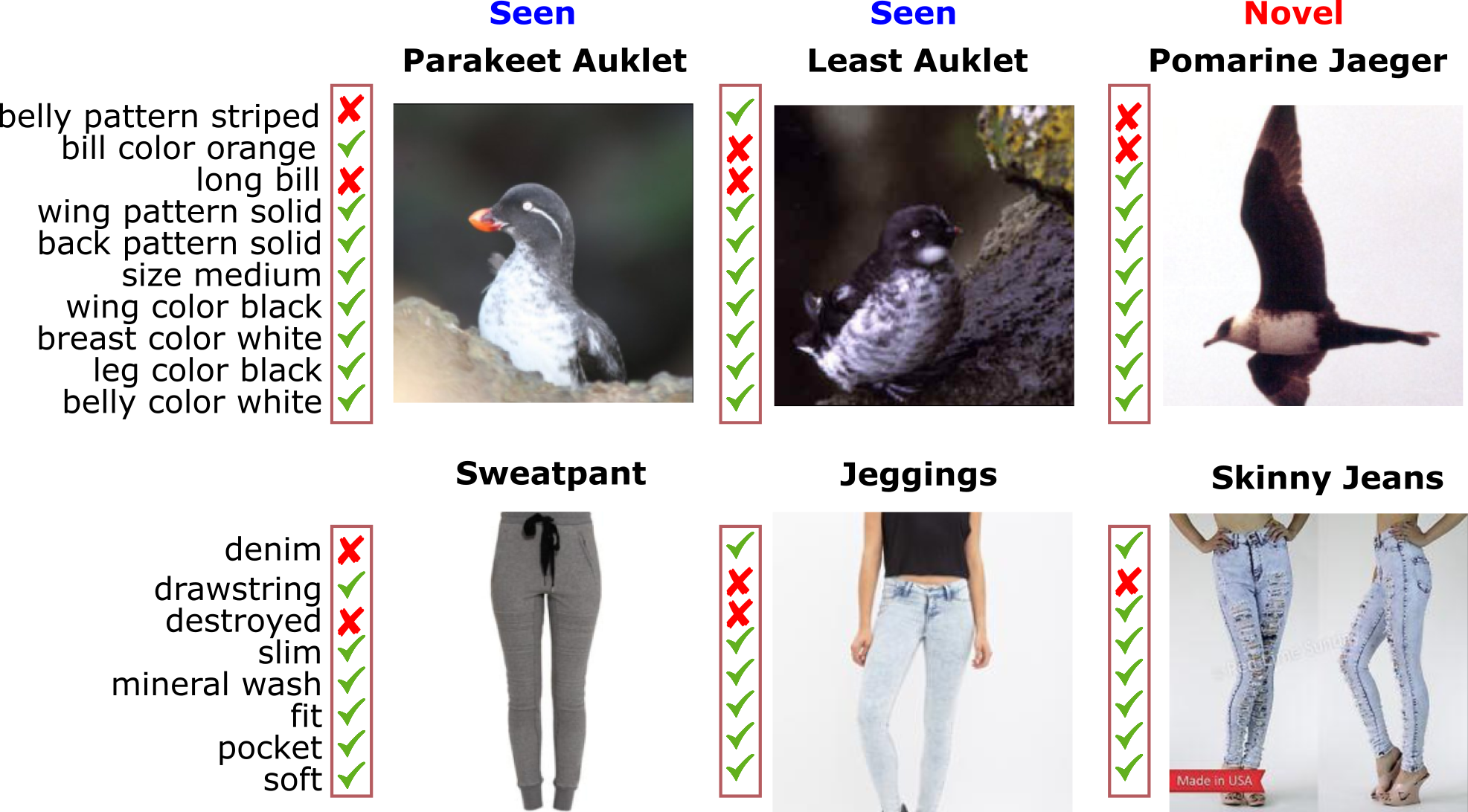}
\caption{
\small{
Three fine-grained classes and their attribute descriptions from the CUB (top) and DeepFashion (bottom) datasets. Notice that these classes are different only in a few attributes.
}
}
\label{fig:att_samples}
\vspace{-2mm}
\end{figure}

Zero/few-shot classification, on the other hand, leverages auxiliary information in the form of class semantic descriptions or few training samples to generalize to classes with no or insufficient samples, respectively. Recent zero-shot works rely on generative models \cite{Felix:ECCV18,Xian:CVPR19,Schonfeld:CVPR19,Zhu:ICCV19,Huang:CVPR19,Sariyildiz:CVPR19} to synthesize features of novel classes to augment the training set.
However, methods based on Generative Adversarial Networks \cite{Huang:CVPR19,Felix:ECCV18,Xian:CVPR19} suffer from low diversity in generated features.
Likelihood-based methods \cite{Arora:CVPR18,Xian:CVPR19,Schonfeld:CVPR19,Zhu:ICCV19,Yu:neurIPS19} promote diversity among generated features, but their generated features are often non-discriminative.
Moreover, these generative methods are not effective in the few-shot learning setting \cite{Vinyals:NIPS16,Snell:NISP17,Finn:ICML17}, where a few samples of novel classes are available. This is because they synthesize novel class features based on only class semantic information, which cannot leverage additional visual information from samples in novel classes to enhance feature generation. While \cite{Hariharan:ICCV17,Wang:18,Li:CVPR20,Chen:CVPR19,Schwartz:NeurIPS18,Zhang:NeurIPS18} are designed to generate features conditioned on visual information, these works ignore class semantic descriptions and cannot synthesize features for classes without any training samples. Although \cite{Xian:CVPR19,Schonfeld:CVPR19} have studied generative models for both zero-shot and few-shot learning, they mainly focus on class semantic information, ignoring few visual samples, or require a large number of unlabeled samples from novel classes in a transductive setting. 

Leveraging the remarkable performance of Convolutional Neural Networks \cite{Simonyan:CORR15,He:CVPR16}, most zero/few-shot works extract image features by pooling local information from image regions into holistic representations \cite{Felix:ECCV18,Xian:CVPR17,Schonfeld:CVPR19,Zhu:ICCV19,Changpinyo:CVPR16,Zhang:CVPR16}.
Although holistic features encode discriminative information among classes, they are not trained to explicitly capture differences among attributes of different classes. Therefore, holistic features cannot describe fine-grained details of novel classes.

It has been shown that fine-grained classes exhibit a compositional structure \cite{Sylvain:ICLR20} in which we only need to recognize basic attributes such as color, shape, or material to recognize a large number of classes expressible in terms of these attributes. 
Few works have explored learning compositional structures.
\cite{Tokmakov:ICCV19,Atzmon:NeurIPS20, Huynh:NeurIPS20} propose to reconstruct image features using attribute features, which either require at least a few samples per class, hence, cannot work in the zero-shot setting, or only work with class semantic vectors, hence, cannot work in the few-shot setting. \cite{Misra:CVPR17,Kato:ECCV18} use neural networks to combine classifiers of simple attribute to recognize novel classes without training samples.
However, these works learn with holistic features which fail to capture attribute details.

\begin{figure}[t]
\centering
\includegraphics[width=0.96\linewidth]{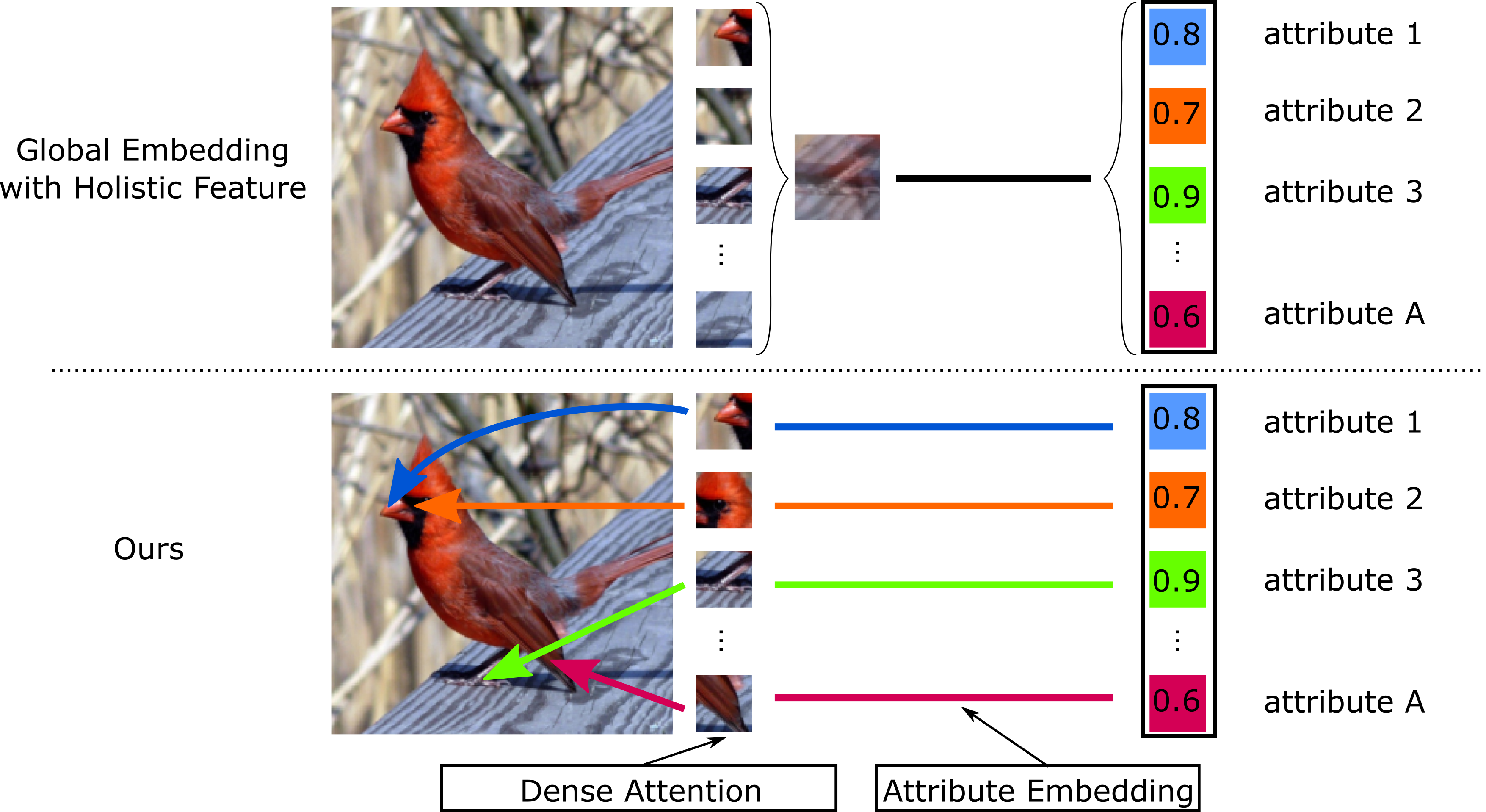}
\vspace{-0mm}
\caption{
\small{Traditional zero-shot classification (top) compresses visual features to perform global embedding with class semantic descriptions, hence, not efficiently capturing fine-grained discriminative visual information. Our method (bottom) finds local discriminative regions through dense attention based on attributes and individually embeds each attribute feature with the attribute semantic description, thus transfers knowledge to novel classes while preserving all fine-grained details.
}
}
\label{fig:dense_attention_motivation}
\vspace{-3mm}
\end{figure}

\subsection{Paper Contributions} 
\begin{figure*} 
\centering
  \includegraphics[width=0.8\linewidth]{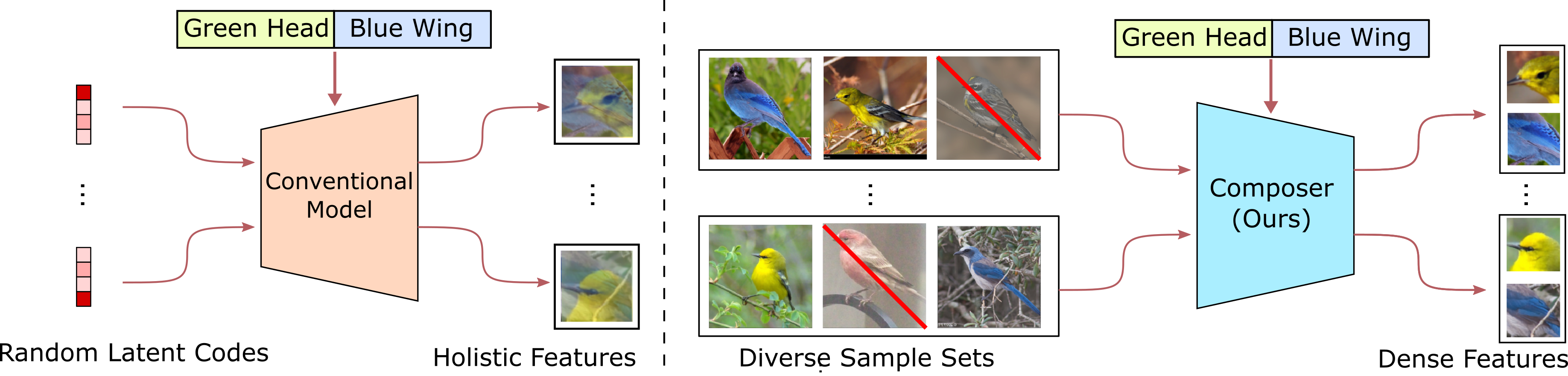}
  \vspace{0mm}
\caption{\small{Left: Traditional generative models synthesize holistic features from random codes lacking fine-grained details. Right: Our compositional model constructs dense features from training samples. By selecting different relevant samples for composition, our method builds diverse features for novel classes.}}
\label{fig:motivation}
\end{figure*}

We develop a novel framework that addresses the limitations of aforementioned methods. 
We propose dense attention and attribute embedding mechanisms capturing all attribute information to distinguish fine-grained differences among classes, as shown in Figure \ref{fig:dense_attention_motivation}.
To improve novel class predictions, instead of generating holistic features, as in Figure \ref{fig:motivation} (left), we learn to extract attribute features from seen classes and recombine them to effectively construct all attribute features of novel classes, see Figure \ref{fig:motivation} (right). 
Our method has several advantages over the state of the art:

\smallskip\noindent\textbf{--} Instead of using holistic features, which lack fine-grained details, we learn to extract and generate dense features consisting of attribute features for hundreds of attributes. 

\smallskip\noindent\textbf{--} Our framework selectively composes features of novel classes from semantically related training samples. It allows specifying different sample sets used for composition that leads to the diversity of composed features. Thus, we control the composition process by constraining the samples that are used to build attribute features of novel classes.

\smallskip\noindent\textbf{--}  Instead of using a generative model to build features and then train a discriminative model, we use a discriminative model to compose features of novel classes in order to train itself. This makes the learning process efficient by removing the need for learning additional generative models.

\smallskip\noindent\textbf{--} Our framework can be naturally adapted to take into account both semantic and visual information of novel classes to improve few-shot performances.

\myparagraph{Paper Organization.}
In Section \ref{sec:related_works}, we discuss about the pros and cons of related works.
Then, we provide an overview on our problem setting as well as our overall approach in Section \ref{sec:overview}.
We propose the dense attention mechanism for extracting fine-grained features for attributes in Section \ref{sec:fine_grained_features}. 
Given these fine-grained features, we introduce an efficient and effective feature composition framework for zero-shot learning and extend it to few-shot learning in Section \ref{sec:generate_fine_grained_features} and \ref{sec:visual_semantic_information}, respectively.
Finally, we demonstrate the effectiveness of our methods on four popular datasets in Section \ref{sec:experiments} and conclude the paper in Section \ref{sec:conclusions}.

\section{Related Works}
\label{sec:related_works}

\myparagraph{Fine-grained Recognition}. The goal of fine-grained recognition is to capture the small but discriminative features across different classes. \cite{Kong:CVPR17,Gao:CVPR16,Lin:ICCV15,Yu:ECCV18} captures the interaction between discriminative feature maps through pooling technique while \cite{Larry:CVPR18,Dubey:ECCV18,Zhang:ICCV19} propose better ways to learn global image features that capture fine-grained details. 
On the other hand, \cite{Zhang:CVPR16,Lin:CVPR15} localize discriminative parts of images through part-based supervision. To avoid the localization annotation of discriminative parts, \cite{Sun:ECCV18,Zheng:ICCV17,Ding:ICCV19,Zheng:CVPR19} localize them in a weakly supervised setting. Despite tremendous success in the fully supervised setting, these works cannot generalize to zero-shot learning, where only high-level attribute descriptions are given for novel classes, or few-shot learning with limited samples in novel classes. 

\myparagraph{Zero-shot Learning}. Early works on zero-shot learning focus on learning joint embedding spaces for visual features and class semantics \cite{Changpinyo:CVPR16,Frome:NIPS13,Norouzi:ICLR14,Xian:CVPR16,Zhang:CVPR16}.
On the other hand, \cite{Lee:CVPR18,Ding:CVPR19,Kampffmeyer:CVPR19,Wang_Gupta:CVPR18} transfer knowledge from seen to novel classes via knowledge graphs which encode richer structures than class semantics.
However, most works on zero-shot learning \cite{Felix:ECCV18,Xian:CVPR19,Schonfeld:CVPR19,Zhu:ICCV19,Huang:CVPR19,Sariyildiz:CVPR19,Changpinyo:CVPR16,Zhang:CVPR16,Xian:CVPR17} rely on holistic image features, which cannot capture local discriminative information from attributes. 
\cite{Yu:NIPS18,elhoseiny:CVPR17} manage to localize distinct visual parts for zero-shot learning despite using costly bounding-box annotations. 
\cite{Huynh-multiatt:CVPR20, Xie:CVPR19,Zhu:NIPS19,Zhu:CVPR19} employ attention mechanism to localize discriminative regions in a weakly-supervised setting, however, they cannot capture every attribute due to the limited number of attention models.
Moreover, the predictions of these methods are often biased towards seen classes due to the lack of training samples for novel classes \cite{Bucher:ICCVW17,Fu:PAMI15}. 
To overcome this issue, \cite{Socher:NIPS13,Atzmon:CVPR18} propose to reduce the probability of seen classes on samples that are out of training distribution, while \cite{Chao:ECCV16} directly calibrates predictions to favor novel classes by adding a fixed margin to prediction scores. Recent work \cite{Huynh-finegrained:CVPR20,Liu:NIPS18} propose calibration losses that consider adjustment of novel class probabilities as training objective functions. However, these works only regularize predictions to prevent seen class bias without transferring knowledge to novel classes.

\myparagraph{Few-shot Learning}. Related to zero-shot learning, few-shot learning aims at recognizing novel classes with few training samples.
To prevent overfitting on these training samples, \cite{Vinyals:NIPS16,Snell:NISP17,Ren:ICLR18} propose to learn suitable metric spaces that generalize to novel classes.
On the other hand, \cite{Ravi:ICLR17,Finn:ICML17,Rusu:ICLR19} frame few-shot learning as finding an optimal model parameters which can quickly adapt to novel classes via fine-tuning.
Recently, \cite{Qi:CVPR18,Qiao:CVPR18} directly predict weights of novel class classifiers while \cite{Xing:NeurIPS19} combines both visual and semantic information to recognize novel classes.
However, most works cannot localize attributes in images for fine-grained recognition.
Although \cite{Tang:CVPR20,Wertheimer:CVPR19,Zhu:IJCAI20} are designed for fine-grained recognition, they either requires strong supervision of attribute locations or cannot learn in zero-shot setting as they ignore semantic information.

\myparagraph{Feature Generation}.
To effectively transfer knowledge to novel classes, recent zero/few-shot learning methods \cite{Bucher:ICCVW17,Felix:ECCV18,Xian:CVPR19,Schonfeld:CVPR19,Hariharan:ICCV17,Wang:18} use generative models to augment a training set with synthesized features of novel classes. 
In zero-shot setting, \cite{Bucher:ICCVW17,Xian:CVPR18} employ Generative Adversarial Networks (GAN) whose generated features often lack diversity due to mode collapse \cite{Bau:ICCV19,Srivastava:neurIPS17,Che:ICLR17} and lack generalization ability due to memorization of training data \cite{Gulrajani:ICLR19,Webster:CVPR19}. On the other hand, \cite{Arora:CVPR18,Xian:CVPR19,Schonfeld:CVPR19,Yu:neurIPS19} propose to use Variational Autoencoders (VAE) that optimize the likelihood of every training sample to enforce diversity. 
However, these methods suffer from posterior collapse \cite{Lucas:neurIPS19,Oord:neurIPS17,Razavi:ICLR19}, which results in generating generic non-discriminative features. To address these issues, \cite{Xian:CVPR19} combines VAE and GAN, while \cite{Zhu:ICCV19} directly learns a feature generator without any encoder or discriminator. However, these improvements are only effective for seen classes with sufficient training samples. Although cycle consistency losses \cite{Felix:ECCV18,Huang:CVPR19,Ni:neurIPS19} directly regulate the generated features of novel classes, they often collapse these features together to align them with semantic vectors of their classes, which results in lack of visual diversity among features.
For few-shot learning, \cite{Hariharan:ICCV17,Wang:18,Li:CVPR20,Chen:CVPR19,Schwartz:NeurIPS18,Zhang:NeurIPS18} propose to generate new poses or viewpoints of novel class samples to capture intra-class variation. 
However, most zero/few-shot works can only synthesize holistic features, which cannot describe fine-grained attributes needed for recognizing novel classes.

\myparagraph{Compositional Learning}. Decomposing concepts into common components is a natural and simple technique for knowledge sharing \cite{Lake:Science15,Lampert:PAMI13,Farhadi:CVPR09,Russakovsky:ECCVW10}.
\cite{Tokmakov:ICCV19,Andreas:ICLR19} learn compositional representations that generalize to classes with few samples. On the other hand, \cite{Chen:CVPR19ImgDeform,Zhang:ICLR18} learn to combine samples for data augmentation. However, these works cannot generalize to novel classes not encountered during training as they require training samples for every class. While \cite{Misra:CVPR17,Kato:ECCV18,Purushwalkam:ICCV19,Yang:CVPR20,Atzmon:NeurIPS20} combine attribute classifiers to recognize novel combination of attributes, they build upon holistic features, which cannot capture fine-grained attribute details. Recently, by examining various zero-shot methods, \cite{Sylvain:ICLR20} shows that zero-shot generalization requires capturing attribute information and their compositional structures.

\myparagraph{}
This work is an extended version of our previous work \cite{Huynh:NeurIPS20,Huynh-finegrained:CVPR20}. In particular, (a) we extend our framework to few-shot setting where both visual and semantic information is used for feature composition. (b) We provide further explanation on the formulation of our framework especially on dense attention mechanism. (c) We report experimental results on few-shot learning to show the effectiveness of our proposed method. (d) We provide additional quantitative and qualitative results to support our design choices on dense attention and attribute grounding.

\section{Overview}
\label{sec:overview}
\subsection{Problem Setting}
We assume that we have two disjoint sets of classes $\C_s$ and $\C_n$, where $\C_s$ denotes seen classes with sufficient training images, $\C_n$ denotes novel classes with limited (less than $m$) or no training samples and $\C \triangleq \C_s \cup \C_n$ denotes the set of all classes. Let $D_s$ and $D_n$ be the set of training samples for seen and novel classes, respectively, and ${(I_1, y_1),\dots,(I_N, y_N)}$ be $N$ training samples, where $I_i$ denotes the $i$-th training image and $y_i \in \C$ corresponds to its class.
Specifically, zero-shot setting requires a model to  recognize novel classes without training samples, $|D_n| = 0$, while the few-shot setting assumes a small amount of training samples in novel classes, i.e., $|D_n| \leq m \ll |D_s|$.
In order to identify novel classes in the zero-shot setting, similar to prior works \cite{Xian:CVPR19,Schonfeld:CVPR19,Zhu:ICCV19}, we assume access to \emph{class semantic} vectors of all classes, $\{\z^c\}_{c\in \C_s \cup \C_n}$, at the training time. More specifically, $\z^c = [z^c_1, \ldots, z^c_A]^\top$, where $z^c_a$ encodes the strength of attribute $a$ in class $c$. We normalize each $\z^c$ to have unit norm to prevent prediction bias toward classes with many attributes. To extract dense features from regions of attribute in images, we also use the \textit{attribute semantic} vectors $\{\v_a\}^A_{a=1}$ as the average of GloVe representation \cite{Pennington:EMNLP14} of words in attribute names. 

\begin{figure*}[t]
\centering
\includegraphics[width=0.98\linewidth]{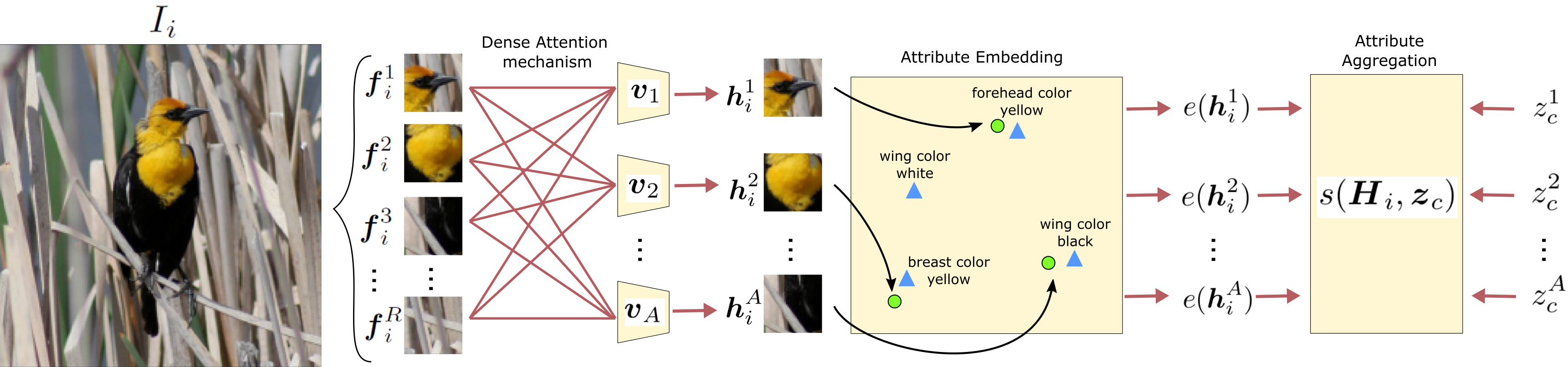}
\vspace{-0mm}
\caption{
\small{
Overview of our dense attention for attributes with attribute embedding. Image features of $R$ regions are extracted and fed into our dense attention mechanism to compute attention features for all attributes. The attention features are then aligned with attribute semantic vectors to measure the scores of attributes in the image, which are combined to form the final prediction.}
}

\label{fig:att_schemantic_fig}
\end{figure*}

\subsection{Fine-Grained Feature Generation for Novel Classes}
We propose to transfer knowledge from seen to novel classes by synthesizing features of novel classes from features of seen classes in order to compensate for the lack of training samples.
Specifically, our goal is to generate features for each attribute and use them to describe novel classes according to their semantic class descriptions, $\z$.
To achieve this, we develop a unified framework consisting of a dense attention mechanism and a compositional feature generation method, which extract attribute features and compose them to synthesize features of novel classes for training.

\section{Learning Fine-Grained Features}
\label{sec:fine_grained_features}
We present a method to extract attribute features for fine-grained recognition. For each attribute, we extract a spatial attention feature from the most relevant regions of an input image.
These attribute features will be subsequently used to compose features of novel classes.
In addition to learning the attention network for attributes, we also fine-tune the attribute semantic vectors to ground textual information into visual domain for more effective transferring knowledge from seen to novel fine-grained classes.

\subsection{Dense Attribute-Based Attention}
The ability to learn visual models of attributes is crucial for transferring knowledge from seen to novel classes. Recent work either embed image features into the class semantic space \cite{Changpinyo:CVPR16,Frome:NIPS13,Norouzi:ICLR14,Xian:CVPR16,Zhang:CVPR16} or generate image features from class semantic vectors \cite{Felix:ECCV18,Xian:CVPR19,Schonfeld:CVPR19,Xian:CVPR18}. However, without localizing each attribute, they ignore discriminative visual features of fine-grained classes, obtaining holistic features that contain information from non-discriminative or irrelevant image regions.

As the first component of our method, we propose an attribute-based spatial attention model, where for each attribute, we localize the most relevant image regions to the attribute to extract an \emph{attribute feature} from a given image.
Recall that $\{\v_a\}_{a = 1}^{A}$ is the set of attribute semantic vectors and $\{ \f_i^r \}_{r=1}^{R} = \text{CNN}_{\theta}(I_i)$ denotes the region features of the image $i$ from a convolutional visual backbone parametrized by $\theta$. For the $a$-th attribute, we define its attention weights of focusing on different regions of image $i$ as,
\begin{equation}
\label{eq:alpha_1}
\alpha(\f_i^r, \v_a) \triangleq \frac{\exp(\v_a^T \W_{\alpha} \f_i^r)}{\sum_{r'} \exp(\v_a^T \W_{\alpha} \f_i^{r'})},
\end{equation}
where $\W_{\alpha}$ denotes a learnable matrix that measures the compatibility between each attribute semantic vector and the visual feature of each region. Using the set of attention weights $\{\alpha(\f_i^r, \v_a)\}_{r=1}^{R}$, we compute the attribute feature for the $a$-th attribute as,
\begin{equation}
\label{eq:visual_attention}
\h^a_i \triangleq \sum_{r=1}^{R} \alpha(\f_i^r, \v_a) \f_i^r.
\end{equation}
Thus, $\h^a_i$ represents the visual feature of the image $i$ that is relevant to the $a$-th attribute according to the attribute semantic vector $\v_a$. Notice that when an attribute is absent in the image, $\h^a_i$ captures the visual evidence used to reject the attribute in the image. For instance, the model could focus on `back belly', and later assigns a negative score to it, to indicate the absence of `white belly', as in Figure \ref{fig:att_samples}.

\begin{figure}[t!]
\centering
\includegraphics[width=0.99\linewidth]{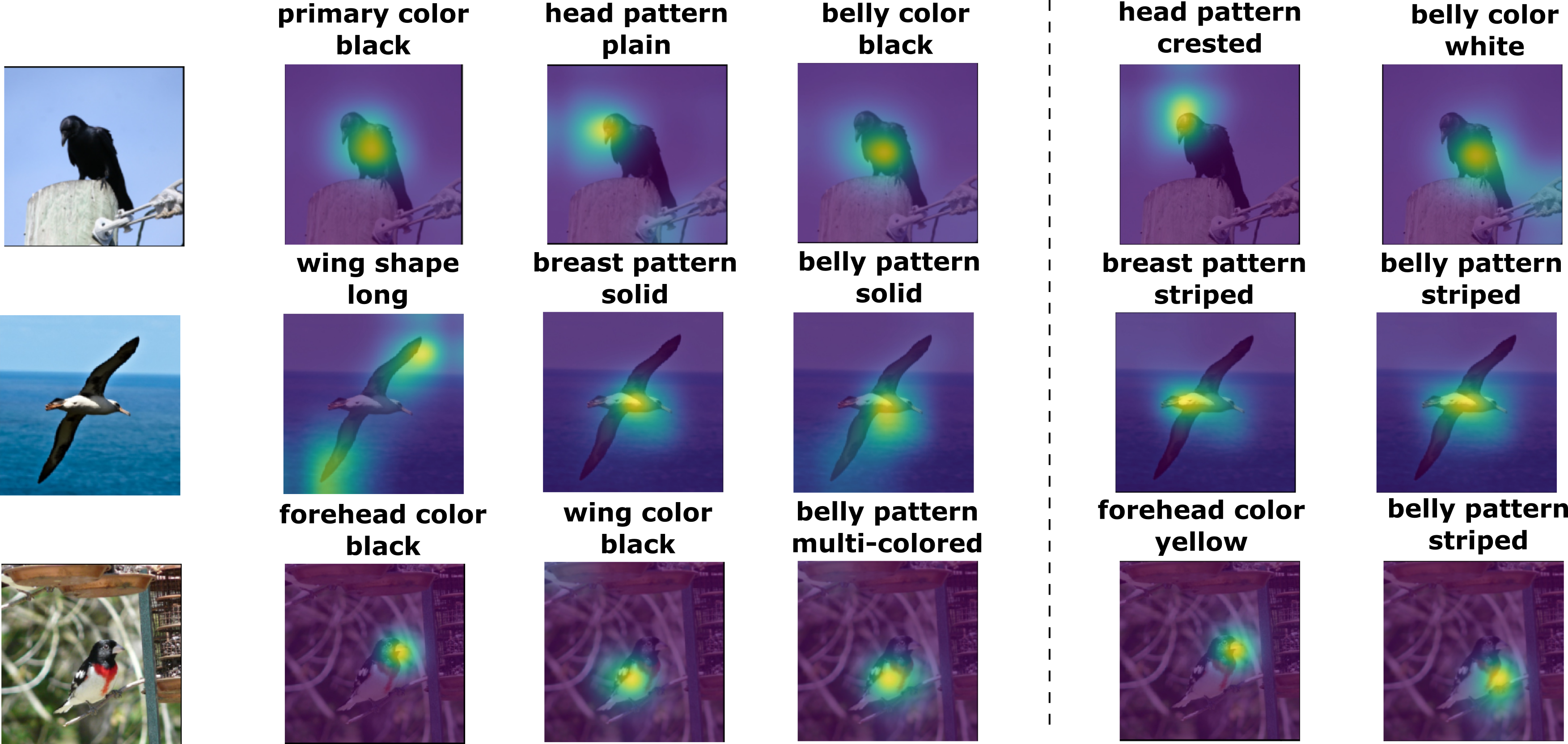}
\vspace{-0mm}
\caption{
\small{
Visualization of attention maps of top positive attribute scores (left) and negative attribute scores (right).
}
}
\vspace{-10pt}
\label{fig:att_samples}
\end{figure}
%

\subsection{Attribute Embedding}
Given the set of attribute features $\{ \h^a_i\}_{a=1}^{A}$ for each training image $i$, our goal is to compute the \emph{class score} of the image $i$ belonging to a class $c$. During training, the class score would be optimized to be large for the ground-truth class $c = y_i$ and small for other classes $c \neq y_i$. To do so, we define $A$ \emph{attribute scores}, where each score measures the strength of having each attribute in the image (recall $A$ is the number of attributes). We fuse these scores using each class semantic vector to find the class score.

More specifically, we define the attribute score $e(\h^a_i)$, which is the confidence of having the $a$-th attribute in the image $i$, by matching the attribute feature $\h^a_i$ with the attribute semantic vector $\v_a$,
\begin{equation}
\label{eq:attr-score}
e(\h^a_i) \triangleq \v_a^T \W_{e} \h_i^a.
\end{equation}
Here, $\W_{e}$ is an embedding matrix that embeds the attribute feature $\h^a_i$ to the $a$-th attribute semantic space. In fact, when the attribute is visually present in an image, the associated image feature would be projected near its attribute semantic vector. We compute the class score as the sum of products between each attribute score, $e(\h_i^a)$, and the strength of having the $a$-th attribute in the $c$-th class, $z^a_c$, as
\begin{equation}
\begin{gathered}
\label{eq:class-score-0}
s(\H_i,\z_c) = \sum_{a=1}^{A} e(\h_i^a) \times z^a_c.
\end{gathered}
\end{equation}
We refer to $\H_i \triangleq \begin{bmatrix} \h^1_{i}, \!\!&\!\!\ldots \!\!&\!\!, \h^A_{i} \end{bmatrix}$ as the \emph{dense feature} matrix, which is the collection of all attribute features for the image $i$. Therefore, when a class $c$ that has attribute $a$ (i.e., $z^a_c > 0$) is present in the image $i$, we learn to maximize the corresponding attribute score $e(\h^a_i)$.

\myparagraph{Loss Function.}
Finally, we train our model to minimize the cross-entropy loss between the predicted and ground-truth labels of training samples,
\begin{equation}
\begin{gathered}
\label{eq:loss_att}
\min_{\W{e},\W_{\alpha},\{\v_a\}^A_{a=1}} - \sum_{i}\log p(y_i|\H_i,\z_{y_i}),\\
p(y_i|\H_i,\z_{y_i}) = \frac{\exp(s(\H_i,\z_{y_i}))}{\sum_{c'}\exp(s(\H_i,\z_{c'}))},
\end{gathered}
\end{equation}
where $p(y_i|\H_i,\z_{y_i})$ is the prediction probability calculated by applying softmax normalization on the score $s(\H_i,\z_{y_i})$. 
\begin{remark}
Notice that in our method, we optimize over attribute semantic vectors $\{\v_a\}_{a=1}^{A}$, which results in visual grounding of each attribute meaning to the visual feature of training images. Also, by sharing $\{\v_a\}_{a=1}^{A}$ among all classes, we effectively allow transferring fine-grained knowledge from seen to novel classes. In the experiments, through ablation studies, we show that fine-tuning the attribute semantic vectors results in significant improvement of the performance.
\end{remark}

Although dense attention and attribute embedding are designed for fine-grained recognition, the entire model is only trained on seen class samples which leads to bias towards seen classes and fails to generalize to novel classes.

\myparagraph{Self-Calibration Approach.}
To overcome this challenge, one approach would to design a calibration loss that allows to shift some of the prediction probabilities from seen to novel classes during training. More specifically, we can define
\begin{equation}
\label{eq:loss_cal}
\mL_{cal} \triangleq -\lambda_{cal}\sum_{i} \log\left(\sum_{n\in \C_{n}} p(n|\H_i,\z_{n}) \right),
\end{equation}
where $\lambda_{cal} \geq 0$ controls the influence of this loss during training.
Thus, minimization of $\mL_{cal}$ in conjunction with the cross-entropy loss, promotes to put nonzero probability on the novel classes during training.
Hence, at testing time, for an image from a novel class, the model can produce a (large) non-zero probability for the true novel class.  

Although this approach can mitigate seen class bias, it simply reduces seen class probabilities without any knowledge transfer between seen and novel classes. Therefore, in the next section, we propose a feature generation framework to synthesize features of novel classes, thus, effectively transferring knowledge from seen to novel classes. As we show in the experiments, our feature generation method significantly improves the performance compared to the self-calibration approach.

\section{Fine-Grained Feature Generation for Zero-Shot Learning}
\label{sec:generate_fine_grained_features}
In this section, we discuss our proposed method for generating dense features, containing all fine-grained features, of novel classes without training samples. 
Our framework first samples a set of candidate feature combinations from which we select the most probable combination having the largest prediction score $p(n|\H,\z_{n})$ for each novel class $n$. The proposed framework alternates between constructing features and updating a discriminative model $p(y|\H,\z)$ by increasing its confidence on the features of novel classes. Algorithm \ref{alg:overall_pipeline} summarizes the steps of our method. We start by defining the compositional property of dense features and propose a framework that generates dense features for novel classes by leveraging this compositional property.

\begin{figure*}[t]
\centering
\includegraphics[width=0.98\linewidth]{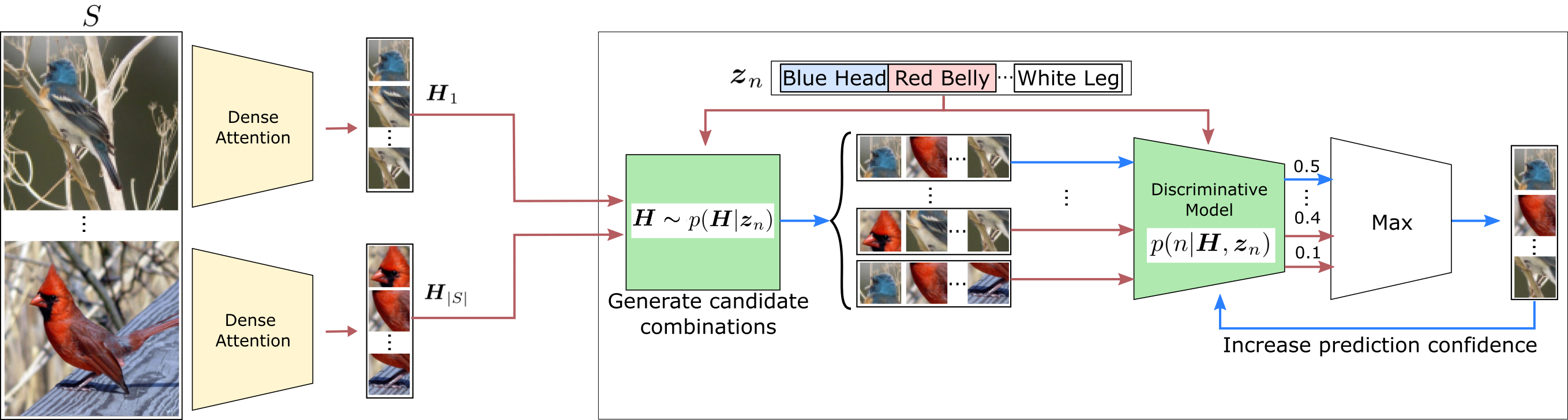}
\vspace{-0mm}
\caption{
\small{
Overview of our compositional zero-shot learning framework. Given a set of samples $S$, we extract dense attribute features $\H_i$ from each sample $i\in S$. For an novel class $u$ with a semantic vector $\z_n$, we generate candidate combinations by sampling from $p(\H|\z_n)$ and use a discriminative model $p(y|\H,\z_n)$ to select the best combination in order to train itself.
}
}

\label{fig:comp_schemantic_fig}
\end{figure*}

\subsection{Dense Feature Generation}
\label{sec:dense_feature_generation}
Since fine-grained classes are different in only a few attributes, we want to generate visual features encoding every attribute information. Thus, we propose to generate the dense feature matrix $\H$, which contains all fine-grained attribute features/information, while discarding irrelevant background information. As such, we use the dense features from training images $\{\H_i\}^N_{i=1}$ as building blocks for our feature composition model.

Our goal is to generate dense features $\H$ for a class $y$ with semantic vector $\z$. Instead of directly learning and maximizing $p(\H | y, \z)$, which is prone to mode/posterior collapse and cannot scale to a large number of attributes and dense features, our insight is to transform a discriminative model $p(y | \H, \z)$, which is easier to train, into a generative model via application of the Bayes rule,
\begin{equation}
\label{eq:comp}
\argmax_{\H}p(\H|y, \z) =\argmax_{\H}p(y|\H, \z)p(\H | \z),
\end{equation}
where the feature prior $p(\H|\z)$ captures the data manifold by assigning high probability to regions containing attribute features and small or zero probability otherwise. Due to the high dimensions of features in $\H$, estimating $p(\H|\z)$ is intractable. To overcome this, we propose a compositional assumption on attribute features where only features $\h_i^a$ from various training samples can be combined into a valid $\H$. This makes the solution-space more tractable since the number of combinations is finite, yet combinatorial. To enforce this assumption, let $S\subset D_s$ be a subset of seen class sample indices from which we construct new features (this is the minibatch used during training). We denote by $\U(S)$ the set of all possible combinations of attribute features from samples in $S$, i.e., 
\begin{equation}
\U(S) \triangleq \big\{ \begin{bmatrix} \h^1_{i_1}, \!&\!\!\ldots ,\!\!&\! \h^A_{i_A} \end{bmatrix} |\ i_a \in S \ \forall a\big\},
\end{equation}
where $\h^a_{i_a}$ is the $a$ attribute feature from sample $i_a$ used to compose $\H$.
We propose to limit the support of the feature prior to only feature combinations in $\U(S)$, i.e., 
\begin{equation}
p(\H|\z) = 0,\ \H \not\in \U(S).
\end{equation}
Selecting features from $\U(S)$ is equivalent to combining attribute features across samples in $S$ to describe novel classes, e.g., if S contains samples of a `blue head, white belly bird' and a `red bird', $\U(S)$ can describe a variety of birds such as a `blue head, red belly bird' or a `red head, white belly bird', see Figure \ref{fig:comp_schemantic_fig}. Thus, we rewrite \eqref{eq:comp} as searching for the best feature composition in $\U(S)$ via maximizing,
\begin{equation}
 \argmax_{\H}p(\H|y,\z) \approx \argmax_{\H\in \U(S)}p(y|\H,\z)p(\H|\z).
\label{eq:compose}
\end{equation}

\begin{remark}
Notice that we use class probability predictions as the criteria for selecting feature combinations. Thus, our framework is robust against missing attributes, where some attributes needed to describe a target class are absent from images in $S$, by resorting to the most probable feature combination  in $\U(S)$.
\end{remark} 

Next, we discuss an efficient way to solve \eqref{eq:compose} via sampling. We propose to train the discriminative model directly on the composed features, leading to an efficient framework that does not need to learn a separate generative model.

\subsection{Efficient Feature Composition}
Since optimizing \eqref{eq:compose} requires an expensive combinatorial search in $\U(S)$, we propose an efficient method to significantly reduce the search space by avoiding irrelevant combinations. To do so, we impose that $\H$ should only be composed from attribute features of \emph{semantically related} samples in $S$ with respect to a target novel class $n$. We define related samples as the ones whose semantic vectors best reconstruct the class semantic vector $\z_n$. We recover the set of related samples in $S$ with respect to a novel class $n$, denoted by $\Q_{n}(S)$, as
\begin{equation}
\begin{split}
\label{eq:Q}
\Q_{n}(S)\triangleq &\argmin_{B\subseteq S} \Big(\min_{\gamma}\|\z_n-\sum_{i\in S'}\z_{i}\gamma_i\|^2_2\Big)\\
&\operatorname{s.t.}~~ |B|\leq k,~~ \gamma_i \geq 0,~ \forall i \in B,
\end{split}
\end{equation}
where $B$ denotes any subset of $S$ of size at most $k$, $\z_i$ denotes the ground-truth class semantic vector of the sample $i$, $\gamma_i$ denotes the reconstruction weight, and $k$ is the number of related samples to select. The nonnegative constraint on the combination weights $\gamma_i$'s ensures that the semantic meaning of each sample does not change due to the reconstruction.
We solve $\eqref{eq:Q}$ using Nonnegative Orthogonal Matching Pursuit \cite{Lin:ICML14}, which greedily adds samples into $\Q_{n}(S)$ to decrease its loss until no further improvement can be made (see the supplementary materials for more details). 

Given the set of related samples $\Q_{n}(S)$, we construct a prior for dense features such that the more related a sample is to the target semantic vector, the more probable its attribute features will be used for composition,
\begin{equation}
\label{eq:attribute_independence}
p(\H|\z_n) \triangleq \prod_{a=1}^{A} p(\h^a_{i_a}|\z_n),
\end{equation}
where we assume the independence of attribute features given the semantic vector.
Notice that attribute dependencies can still be captured via classifier predictions $p(y|\H,\z)$ in \eqref{eq:compose} for feature generation.
We propose to compute probability of using attribute features from training samples as,
\begin{equation}
p(\h^a_{i_a}|\z_n)\triangleq \begin{cases}
      \frac{\exp\big(\z_{i_a}^\top\z_n \big)^\beta}{\sum_{i\in \Q_{n}(S)}\exp\big(\z_{i}^\top\z_n\big)^\beta }, & \text{if}\ i_a \in \Q_{n}(S), \\
      0, & \text{otherwise,}
    \end{cases}
\end{equation}
where $\z^{i_a}$ is the ground-truth class semantic vector of sample $i_a$ used to compose attribute $a$, $\beta$ is a non-negative scalar that controls the probability of using attribute features from related samples. When $\beta \gg 0$, the prior would mostly include attribute features from the most related sample and  when $\beta=0$, it would uniformly sample attribute features from all related samples. Notice that we measure the relatedness as the cosine similarity between the sample semantic $\z_{i_a}$ and the target semantic $\z$.
The prior also assigns zero probability to combinations of samples outside the related sample set $\Q_{n}(S)$ to exclude these combinations.
Indeed, the prior allows us to sample a set of candidate features to find the most probable feature, thus we avoid searching through all combinations in $\U(S)$. Specifically, we first samples a set of candidate combinations $M_n(S)$ from which we seek the combination that maximizes the product of an novel class probability and the prior, 
\begin{equation}
\begin{split}
H_n(S) &\triangleq \argmax_{\H\in M_n(S)}p(n|\H,\z_n)p(\H|\z_n),\label{eq:novel_compose}\\
M_n(S) &\triangleq \{\H|\H\thicksim p(\H|\z_n)\},
\end{split}
\end{equation}
where $H_n(S)$ is the most probable dense feature of class $n$ and $M_n(S)$ is constructed by sampling $b$ candidate combinations according to $p(\H|\z_n)$\footnote{Notice that a class can be described by different semantic vectors which reflects its visual variations. For simplicity, we use a single semantic vector per class.}.

\myparagraph{Loss Function.}
Having the composed features of novel classes $\{H_n(S)\}_{n\in \C_n}$, we train the discriminative model by increasing its prediction confidence on these features as novel classes while maintaining the confidence on seen class samples via the following cross-entropy loss,
\begin{equation}
\begin{split}
\min_{\W{e},\{\v_a\}^A_{a=1}} \E_{S} \Big[-\frac{1}{|S|} \sum_{i\in S} \log p\big(y_i|\H_i,\z_{y_i} \big) \\
- \frac{1}{|\C_n|}\sum_{u\in \C_n} \log p\big(n|H_n(S),\z_n\big) \Big].
\label{eq:self-compose}
\end{split}
\end{equation}
Here, we transfer knowledge from seen to novel classes by recognizing novel combinations of features as novel classes.
Thus, the discriminative model learns the existence of novel classes to avoid seen class bias.
We alternate between composing features \eqref{eq:novel_compose} and minimizing the loss \eqref{eq:self-compose} on a random sample set $S$ in each iteration until convergence.
By randomizing the sample set $S$, we effectively ensure the diversity among composed features by enforcing them to be built from various sets of samples.

For inference, we recognize a test image by finding the most probable class according to the discriminative model: $c^* = \argmax_{c \in \C_s \cup \C_n} p(c|\H,\z_c)$.

\begin{algorithm}[H]
\small
\caption{Compositional Feature Generation}
\label{alg:overall_pipeline}
\begin{algorithmic}
\State \textbf{Input:} Training set $D_s,D_n$, randomly initialized $\alpha(\f,\v),p(y|\H,\z)$
\State \texttt{\textbackslash\textbackslash 1st Stage: Learn Dense Attention Model}
\For{$t = 1, \ldots, N_{att}$}
\State Train Dense Attention Model $\alpha(\f,\v)$ via \eqref{eq:loss_att} 
\EndFor
\State \texttt{\textbackslash\textbackslash 2nd Stage: Compose Dense Feature}
\For{$t = 1, \ldots, N_{comp}$}
\State Sample $S \subset D_s, S_n \subset D_n$
\State Extract dense features $\{\H_{i}\}_{i\in S \cup S_n}$ with $\alpha(\f,\v)$ via \eqref{eq:visual_attention}
\State Generate features of novel classes via \eqref{eq:novel_compose} or \eqref{eq:novel_compose_fs}
\State Fine-tune $p(y|\H,\z)$ via \eqref{eq:self-compose} or \eqref{eq:self-compose_fs}
\EndFor
\State \textbf{Output:} Optimal parameters
\end{algorithmic}
\end{algorithm}

\section{Fine-Grained Feature Generation for Few-Shot Learning}
\label{sec:visual_semantic_information}
In this section, we propose to extend our framework for few-shot learning where there are limited training samples of novel classes.
Let $\H_j$ and $\z_j$ are the dense feature and semantic vector, respectively, of a training sample of novel class $j \in D_n$.
Our goal is to generate features for the novel class $y_j$ that takes into account this class training samples by conditioning the feature prior on both semantic and visual information.
As such, we introduce a novel \textit{visual-semantic feature} based on both $\H_j$ and $\z_j$ as,
\begin{equation}
\bar{\h}_j = \sum^A_{a=1}\h^a_j z^a_j,
\end{equation}
where $\bar{\h}_j$ is the visual-semantic feature of sample $j$.
Here, we use the semantic information in $\z_j$ to highlight the discriminative attribute features in $\H_j$, that represent the novel class $y_j$.
If attribute $a$ is indicative for class $y_j$ , i.e., $\z^a_j >> 0$, then $\bar{\h}_j$ would mostly include this attribute feature $\h^a_j$ and would suppress other irrelevant attribute features with $z^a_j=0$.
We propose to synthesize features of few-shot classes by conditioning the generation process on their visual-semantic features as,
\begin{multline}
\argmax_{\H}p(\H|y_j,\z_j,\bar{\h}_j) \\
 \approx \argmax_{\H\in \U(S)}p(y_j|\H,\z_j)p(\H|,\bar{\h}_j).
\label{eq:compose_fs}
\end{multline}
To efficiently solve the generation process, we employ a similar sampling technique as in the zero-shot setting which first samples candidates, $M_j(S)$, from the prior distribution and selects the best feature combination among them,
\begin{equation}
\begin{split}
H_j(S) &\triangleq \argmax_{\H\in M_j(S)}p(y_j|\H,\z_j)p(\H|\bar{\h}_j),\\
\label{eq:novel_compose_fs}
M_j(S) &\triangleq \{\H|\H\thicksim p(\H|\bar{\h}_j)\}.
\end{split}
\end{equation}
To compute the prior distribution, we identify a set of visually and semantically related samples $\Q_j(S)$ which best reconstruct the visual-semantic feature of $j$, 
\begin{equation}
\begin{split}
\Q_j(S) \triangleq &\argmin_{B\subseteq S} \Big(\min_{\gamma}\| \bar{\h}_j -\sum_{i\in B}\bar{\h}_i \gamma_i\|^2_2 \Big)\\ 
&\operatorname{s.t.}~~ |B|\leq k,~~ \gamma_i \geq 0,~ \forall i \in B.
\end{split}
\end{equation}
Given the related sample set, we simplify the prior via attribute-independence assumption $p(\H|\bar{\h}_j) \triangleq \prod_{a=1}^{A} p(\h^a_{i_a}|\bar{\h}_j)$ similar to \eqref{eq:attribute_independence}.
We use the cosine similarity of visual-semantic features to express the probability of using sample $i_a$ to represent sample $j$,
\begin{equation}
p(\h^a_{i_a}|\bar{\h}_j)\triangleq \begin{cases}
      \frac{\exp\big(\bar{\h}_{i_a}^\top\bar{\h}_j \big)^\beta}{\sum_{i\in \Q_{j}(S)}\exp\big(\bar{\h}_{i}^\top\bar{\h}_j\big)^\beta }, & \text{if}\ i_a \in \Q_{u}(S). \\
      0, & \text{otherwise.}
    \end{cases}
\end{equation}
By using visual-semantic features, we effectively incorporate both visual and semantic information for feature generation while leveraging our efficient composition procedure.

\myparagraph{Loss Function.}
Unlike zero-shot setting, where a model can only be trained on synthesized features of novel classes, few-shot setting provides limited novel class samples for training.
Thus, we propose to optimize the loss function with respect to both real and synthesized features of novel classes as:  
\begin{equation}
\begin{split}
\min_{\W{e},\{\v_a\}^A_{a=1}} \E_{S,S_n} \Big[-\frac{1}{|S|} &\sum_{i\in S} \log p\big(y_i|\H_i,\z_{y_i} \big) \\
-\frac{1}{|S_n|(1+\lambda)} &\sum_{j\in S_n} \log p\big(y_j|\H_j,\z_j \big)\\
-\frac{\lambda}{|S_n|(1+\lambda)} &\sum_{j\in S_n} \log p\big(y_j|H_j(S),\z_j \big) \Big],
\label{eq:self-compose_fs}
\end{split}
\end{equation}
where $S \subseteq D_s, S_n \subseteq D_n$ are sets of samples from seen and novel classes, respectively, and $\lambda \geq 0$ trades off the influences of learning from real or synthesized features in novel classes during training.

\section{Experiments}
\label{sec:experiments}
We demonstrate the effectiveness of our framework, referred to as \emph{Composer}, on four popular datasets: DeepFashion \cite{Liu:CVPR16}, AWA2 \cite{Xian:PAMI18}, CUB \cite{Welinder:report10}, SUN \cite{Patterson:CVPR12}. 
We first discuss the datasets, evaluation metrics, implementation details and baselines. 
We then present the zero-shot and generalized zero-shot performances in pre-trained and fine-tuned feature extractor settings. We show comparisons between compositional models and current generative models, effects of hyper-parameters, and ablation studies on dense attention as well as feature composition.
Finally, we demonstrate the effectiveness of our framework on few-shot settings and the effect of learning from real and synthesized features.  
\subsection{Experimental Setup}
\myparagraph{Datasets:}
We conduct experiments on visual recognition datasets: DeepFashion \cite{Liu:CVPR16}, AWA2 \cite{Xian:PAMI18}, CUB \cite{Welinder:report10}, and SUN \cite{Patterson:CVPR12} having different data statistics, as shown in Figure \ref{tab:data_stats}.
DeepFashion \cite{Liu:CVPR16} contains images of fine-grained clothing categories with 36 seen classes and 10 novel classes.
Due to attribute redundancy in DeepFashion, we only select top 300 most discriminative attributes by choosing attributes having low class entropy.
AWA2 \cite{Xian:PAMI18} consists of animal images with 40 seen classes and 10 novel classes described by 85 attributes.
CUB \cite{Welinder:report10} is a fine-grained bird dataset with 47 images per class for 150 seen classes and 50 novel classes. Each class is carefully annotated with 312 attributes which can be grounded in images.
Finally, SUN \cite{Patterson:CVPR12} is a visual scene dataset with 645 seen classes and 72 novel classes with only 16 images per class. 
We follow the data splits of \cite{Xian:PAMI18} for AWA2, CUB, and SUN.
On DeepFashion, we partition the categories into 36 seen and 10 novel classes, in order to have a sufficient number of training samples in novel classes.
We use the original training/testing split of the dataset to further divide seen classes into training and testing sets.

\begin{table}[t]
\centering
\resizebox{1\linewidth}{!}{
\begin{tabular}{|c|c|c|c|}
\hline
Dataset & \# attributes & \makecell{\# seen (val) \\ /\ novel classes} & \makecell{\# training \\/\ testing samples}\\
\hline\hline
DeepFashion  & {300} & 30 (6) /\ 10 & \textbf{204,885} /\ \textbf{84,337}\\
AWA2  & 85 & 27 (13) /\ 10 & 23,527 /\ 13,795\\
CUB  & \textbf{312} & 100 (50) /\ 50 & 7,057 /\ 4,731\\
SUN  & 102 & \textbf{580} (\textbf{65}) /\ \textbf{72} & 10,320 /\ 4,020\\
\hline
\end{tabular}
}
\caption{\small{Statistics of the datasets used in our experiments.}}
\label{tab:data_stats}
\vspace{-5mm}
\end{table}

{
\setlength{\tabcolsep}{4pt}
\renewcommand{\arraystretch}{1.2} 
\begin{table*}[t]
 \centering
 \resizebox{\linewidth}{!}{%
   \begin{tabular}{|c| c |c c c |c | c c c |c | c c c | c | c c c  |}
   \hline
    \multirow{2}{*}{Method}& \multicolumn{4}{c|}{\textbf{DFashion} (5691 images/class)} & \multicolumn{4}{c|}{\textbf{AWA2} (588 images/class)} &  \multicolumn{4}{c|}{\textbf{CUB} (47 images/class)} & \multicolumn{4}{c|}{\textbf{SUN} (16 images/class)} \\
    \cline{ 2- 17}
     & \color{blue}$n\rightarrow n$ & $a\rightarrow s$ & $a\rightarrow n$ & $H$ & \color{blue}$n\rightarrow n$ & $a\rightarrow s$ & $a\rightarrow n$ & $H$  & \color{blue}$n\rightarrow n$ & $a\rightarrow s$ & $a\rightarrow n$ & $H$  & \color{blue}$n\rightarrow n$ & $a\rightarrow s$ & $a\rightarrow n$ & $H$   \\
     \hline
     \multicolumn{17}{|c|}{Pre-trained Setting} \\
     \hline
\texttt{\makecell{MLSE}}~\cite{Ding:CVPR19} &  -	&-&	-&	-&	67.8&	83.2&	23.8&	37.0&	64.2&	71.6&	22.3&	34.0&	62.8&	36.4&	20.7& 26.4\\ 
\texttt{\makecell{CVC}}~\cite{Li:ICCV19} &  -	&-&	-&	-&	71.1&	81.4&	56.4&	66.7&	54.4&	47.6&	47.4&	47.5&	62.6&	36.3&	42.8& 39.3\\ 
\texttt{\makecell{TripletLoss}}~\cite{Cacheux:ICCV19} &  -	&-&	-&	-&	67.9&	\textbf{83.2}&	48.5&	61.3&	63.8&	52.3&	55.8&	53.0&	63.8&	30.4&	47.9& 36.8\\  
     \texttt{\makecell{f-VAEGAN-d2}}~\cite{Xian:CVPR19} &  -	&-&	-&	-&	71.1&	70.6&	57.6&	63.5&	61.0*&	60.1*&	48.4*&	53.6*&	\color{blue}\textbf{64.7}&	\textbf{38.0}&	45.1& \textbf{41.3}\\  
\texttt{\makecell{CADA-VAE}}~\cite{Schonfeld:CVPR19} & - & - & - &	- & - &	75.0&	55.8&	64.0&	- &	53.5&	51.6&	52.5&	-	&35.7&	47.2&	40.7\\
\texttt{\makecell{f-Translator}}~\cite{Zhu:ICCV19} &  40.7	&30.5&	23.9&	26.8&	70.4&	72.6&	55.3&	62.6&	58.5&	54.8&	47.0&	50.6&	61.5&	36.8&	45.3& 40.6\\ 
\texttt{\makecell{Self-Calibration}} &  38.7&	\textbf{35.7}&	21.6&	27.0&	67.7&	75.3&	60.5&	67.1&	66.3&	\textbf{59.6}&	56.7&	58.1&	60.0&	23.2&	52.0&	32.1 \\
\texttt{Composer (Ours)} & \color{blue}\textbf{43.0}&	32.9&	\textbf{31.2}&	\textbf{32.0}&	\color{blue}\textbf{71.5}&	77.3&	\textbf{62.1}&\textbf{68.8}&	\color{blue}\textbf{69.4}&	56.4&	\textbf{63.8}&	\textbf{59.9}&	62.6&	22.0&	\textbf{55.1}&	31.4\\
     \hline
     \multicolumn{17}{|c|}{Fine-tuned Setting} \\
     \hline
\texttt{\makecell{SMA}}~\cite{Yu:NIPS18} & - &-&	-&	-& 68.8	& 87.1 &	37.6 & 52.5 &	71.0 & 71.3 & 36.7	&	48.5 & - & - &	-& -\\
\texttt{\makecell{LFGAA+SA}}~\cite{Liu:ICCV19} & - &-&	-&	-& 68.1	& \textbf{90.3} &	50.0 & 64.4 &	67.6 & \textbf{79.6} & 43.3	&	64.4 & 61.5 & 34.9 &	20.8& 26.1\\
\texttt{\makecell{APN}}~\cite{Xu:NeurIPS20} & - & - & - & - & 68.4	&	78.0 &	56.5&	65.5 &	72.0 &	69.3&	65.3&	\textbf{67.2}& 61.6	&34.0&	41.9&	37.6\\
\texttt{\makecell{f-VAEGAN-d2}}~\cite{Xian:CVPR19} &  -	&-&	-&	-&	70.3&	76.1&	57.1&	65.2&	72.9*&	75.6*&	63.2*&	68.9*&	\color{blue}\textbf{65.6}&	\textbf{37.8}&	50.1& \textbf{43.1}\\  
\texttt{\makecell{AREN+CS}}~\cite{Xie:CVPR19} & 41.0 & 36.3 & 27.5 & 31.3 & 67.9	&	79.1&	54.7&	64.7&	71.8 &	69.0&	63.2&	{66.0}& 60.6	&32.3&	40.3&	35.9\\
\texttt{\makecell{Self-Calibration}} &  44.1 & 41.3 &	26.5 & 32.3 &	66.7 & 72.1 & 61.7 & 66.5 & 69.7 & 55.4 & 64.1 & 59.4 & 59.5 &	25.0 &	51.5 &	33.7 \\ 
\texttt{Composer (Ours)} &\color{blue}\textbf{47.3}&	\textbf{42.3} &	\textbf{32.8} & \textbf{36.9}& \color{blue}\textbf{75.4} &	76.1&	\textbf{62.2}&	\textbf{68.5}&	\color{blue}\textbf{74.0}&	61.6&	\textbf{66.3}&	63.9&61.0&	24.7&	\textbf{53.4}&	33.8\\
     \hline
\end{tabular}
   } 
   \caption{\small{Performances on DeepFashion, AWA2, CUB and SUN. We report zero-shot accuracy ($n\rightarrow n$) in the zero-shot setting and seen class accuracy ($a\rightarrow s$), novel class accuracy ($a\rightarrow n$), harmonic mean ($H$) in generalized zero-shot setting. * indicates the usage of extra supervision from human captions.}
}

\label{tab:gzsl}
\end{table*}
}

\begin{figure*} 
\centering
\begin{minipage}[b]{0.3\linewidth}
  \includegraphics[width=1\textwidth]{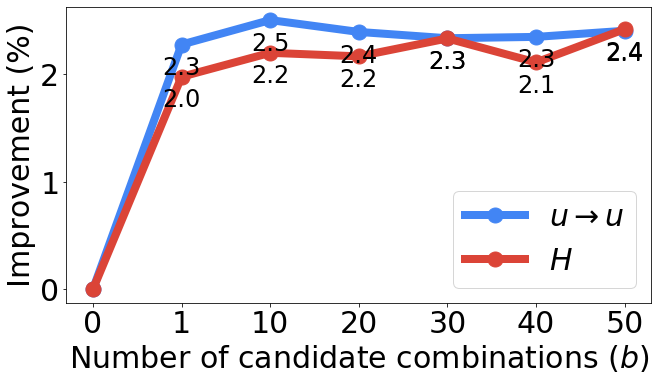}
  \vspace{-1mm}
\end{minipage}
\hspace{2mm}
\begin{minipage}[b]{0.3\linewidth}
  \includegraphics[width=1\textwidth]{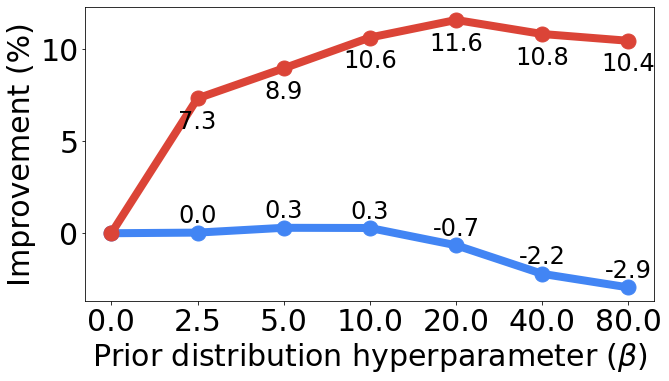}
    \vspace{-5mm}
\vspace{4mm}
\end{minipage}
\hspace{2mm}
\begin{minipage}[b]{0.3\linewidth}
  \includegraphics[width=1\textwidth]{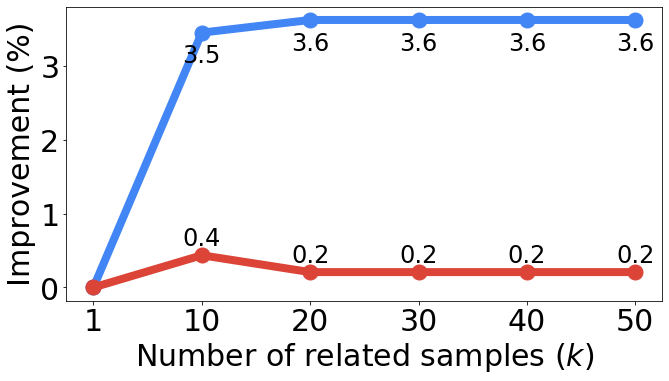}
    \vspace{-5mm}
\vspace{4mm}
\end{minipage}
\vspace{-3.5mm}
\caption{\small{Effects of hyperparameters on harmonic mean ($H$) and zero-shot accuracy ($n\rightarrow n$) on DeepFashion in fine-tuned setting.}}
\label{fig:DeepFashion_hyper_params}
\end{figure*}

\myparagraph{Evaluation Metrics:}
Similar to \cite{Xian:PAMI18,Schonfeld:CVPR19}, we measure the top-1 accuracy on two settings: i) zero/few-shot learning ($n\rightarrow n$) where testing images are from novel classes thus the model only need to distinguish among novel classes and ii) generalized zero/few-shot learning where testing images comes from both seen and novel classes. In the latter setting, we measure the accuracy when recognizing seen classes ($a\rightarrow s$) and novel classes ($a\rightarrow n$). We compute the harmonic mean ($H$) between seen and novel class accuracy to measure the trade-off between these performances.
In few-shot learning, we report the mean performances over 10 runs with different random seeds following \cite{Schonfeld:CVPR19}.

\myparagraph{Baselines:} 
For zero-shot learning, we compare our method with 3 main approaches: generating features of novel classes, learning transferable visual representations, and learning compatibility functions. \texttt{f-VAEGAN-d2} \cite{Xian:CVPR19} and \texttt{CADA-VAE} \cite{Schonfeld:CVPR19} generate features of novel classes using GANs or VAEs while \texttt{f-Translator} \cite{Zhu:ICCV19} directly optimize training data likelihood to learn a feature generator. On the other hand, \texttt{CVC} \cite{Li:ICCV19} generates classifiers of novel classes.
To learn visual representation, \texttt{SMA} \cite{Yu:NIPS18}, \texttt{AREN+CS} \cite{Xie:CVPR19} and \texttt{APN} \cite{Xu:NeurIPS20} use attention models to capture discriminative class features while \texttt{MLSE} \cite{Ding:CVPR19} learns latent class representations via semantic graphs.
Finally, \texttt{TripletLoss} \cite{Cacheux:ICCV19} learns a compatibility function between visual and semantic information by accounting for the semantic similarity among classes. \texttt{LFGAA+SA} \cite{Liu:ICCV19} proposes a dynamic compatibility function that adapts to attributes appearing in an image.
To show the effectiveness of feature composition, we report performances of our \texttt{Self-Calibration} approach, which uses the calibration loss \eqref{eq:loss_cal} with $\lambda_{cal}=0.1$, which achieves the best performances on all datasets. 
On DeepFashion, we run each baseline using their released codes with their default settings. On the remaining datasets, we use the performances reported in their papers to ensure their best performances.
For few-shot learning, we compare with two discriminative methods: \texttt{WeightImprint} \cite{Qi:CVPR18}, which predicts classifiers for novel classes from few samples, \texttt{AM3} \cite{Xing:NeurIPS19}, which leverages both semantic and visual information to recognize novel classes.
Moreover, we report \texttt{CADA} as the state-of-the-art generative model for few-shot learning.  

\myparagraph{Implementation Details:} 
Following \cite{Xian:CVPR17}, we resize images to $224\times 224$ and extract features using the ResNet101 backbone \cite{He:CVPR16} for our method. Our setting is comparable with the above baselines except for \texttt{SMA} using VGG19 and \texttt{LFGAA} combining VGG19, GoogleNet, and ResNet101.
We use the feature map of the last convolutional layer whose size is $7\times7\times2048$ and treat it as features from $7\times7=49$ regions. 
We implement our framework in PyTorch and optimize it using RMSprop\cite{Tijmen:COURSERA12} with the default setting, learning rate of $0.0001$ and batch size of $50$ having an equal number of samples per class. We train the dense attention model on seen classes and use it to compose dense features for at most $N_{att}=2000$ and $N_{comp}=4000$ iterations, respectively, on a NVIDIA V100 GPU. 
To prevent seen class bias in zero-shot setting, we add a margin of $1$ to novel class scores and $-1$ to seen class scores, which reduces the dominance of seen classes similar to \cite{Chao:ECCV16}.  
We experiment in two settings: i) using pre-trained ImageNet features (pre-trained setting) and ii) fine-tuning the ResNet backbone on each dataset (fine-tuned setting). 
To measure the robustness of our method, we fix the hyperparameters at $\beta=5,k=5,b=50$ ($\beta=10,k=10,b=50$) for the pretrained (fine-tuned) setting on all datasets.
For few-shot learning, we set $\lambda=0.5$ and use a pre-trained ResNet101 backbone in all experiments for fair comparison with \cite{Schonfeld:CVPR19}.

\subsection{Zero-Shot Experiments}

\begin{table*}
\begin{minipage}{0.6\textwidth}
\centering
\resizebox{1\textwidth}{!}{
   \begin{tabular}{|c|c | c c c |c | c c c |}
   \hline
    \multirow{2}{*}{Method} & \multicolumn{4}{c|}{\textbf{AWA2}} &  \multicolumn{4}{c|}{\textbf{CUB}}\\
    \cline{ 2- 9}
     & \color{blue}$n\rightarrow n$ & $a\rightarrow s$ & $a\rightarrow n$ & $H$  & \color{blue}$n\rightarrow n$ & $a\rightarrow s$ & $a\rightarrow n$ & $H$ \\
     \hline
     \multicolumn{9}{|c|}{Generative models} \\
     \hline
\texttt{\makecell{f-Translator}}~\cite{Zhu:ICCV19} &	70.4&	72.6&	55.3&	62.6&	58.5&	54.8&	47.0&	50.6\\ 
     \texttt{\makecell{f-VAEGAN-d2}}~\cite{Xian:CVPR19} &	71.1&	70.6&	57.6&	63.5&	61.0&	60.1&	48.4&	53.6\\  
\texttt{\makecell{CADA-VAE}}~\cite{Schonfeld:CVPR19} &-&	75.0&	55.8&	64.0&	- &	53.5&	51.6&	52.5\\
\texttt{\makecell{Attribute GANs}} &65.1&	75.2&	58.1&	65.6&	- &	-&	-&-\\
\hline
     \multicolumn{9}{|c|}{Compositional models} \\
     \hline
     \texttt{\makecell{Random Comp}}&	65.5&	76.7&	56.6&	65.1&	67.3&	\textbf{64.4}&	51.2&	57.0\\ 
    \texttt{Composer (Ours)} &	\color{blue}\textbf{71.5}&	\textbf{77.3}&	\textbf{62.1}&	\textbf{68.8}&	\color{blue}\textbf{69.4}&	56.4&	\textbf{63.8}&	\textbf{59.9}\\
     \hline
\end{tabular}
}
\end{minipage}
\hfill
\begin{minipage}{0.4\textwidth}
\resizebox{1\textwidth}{!}{
   \begin{tabular}{|c| c |c c c |}
   \hline
	\multirow{2}{*}{Method} & \multicolumn{4}{c|}{\textbf{DeepFashion}}\\
    \cline{ 2- 5}
     & \color{blue}$u\rightarrow u$ & $a\rightarrow s$ & $a\rightarrow u$ & $H$ \\
     \hline
      \makecell{\texttt{No Comp}}& 44.6 & 54.2 & 2.6 & 5.0\\
	\hline
 	\makecell{\texttt{Random Comp}}& 44.9&  40.4	& 30.1 & 34.5 \\
 	\hline
	\makecell{$p(\H|\z)$\\\texttt{Comp}}& 43.9& 36.9 & \textbf{37.1} & 36.5 \\
 	\hline
	\makecell{$p(y|\H,\z)p(\H|\z)$\\\texttt{Comp} (fixed $S$)}& 46.9 & \textbf{44.7} & 26.9 & 33.6 \\
	\hline
	\makecell{\texttt{Composer (Ours)}}&	\color{blue}\textbf{47.3}&	42.3 &	32.8 & \textbf{36.9} \\
	\hline
\end{tabular}
}
\end{minipage}
\caption{\small{Left: Comparison between generative models and compositional models on AWA2 and CUB in the pre-trained setting. Right: Ablation study on DeepFashion in the fine-tuned setting.}}
\vspace{0mm}
\label{tab:composition_strategies_DeepFashion_ablation}
\end{table*}

\myparagraph{Zero-Shot Learning:}
Table \ref{tab:gzsl} shows the zero-shot accuracy ($n\rightarrow n$) across different datasets. In the pre-trained setting, our method significantly outperforms other methods by at least 2.3\% and 3.1\% on DeepFashion and CUB, respectively while having comparable performance with the state-of-the-art method on AWA2.
In the fine-tuned setting, we improve at least 3.2\%, 5.1\%, and 2.2\% on DeepFashion, AWA2, and CUB respectively. Although we do not uses human captions as extra supervision in CUB, our method significantly surpasses \texttt{f-VAEGAN-d2} by 8.4\% (1.1\%) on pre-trained (fine-tuned) settings.
Our strong performances demonstrate that dense feature composition can effectively describe fine-grained attribute details of novel classes.
Having only 16 samples per class in SUN does not allow to effectively train dense attention model, which results in low performances.

\begin{figure*} 
\centering
\begin{minipage}[b]{0.24\linewidth}
  \includegraphics[width=1\textwidth]{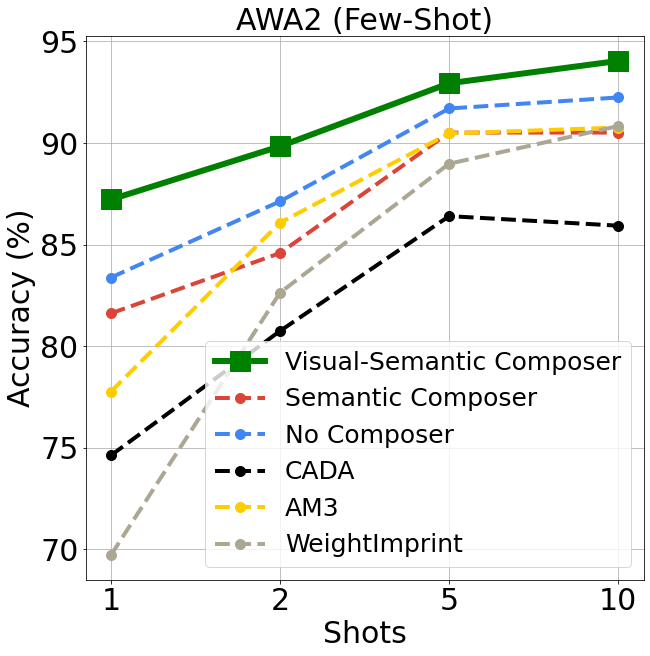}
  \vspace{-1mm}
\end{minipage}
\begin{minipage}[b]{0.24\linewidth}
  \includegraphics[width=1\textwidth]{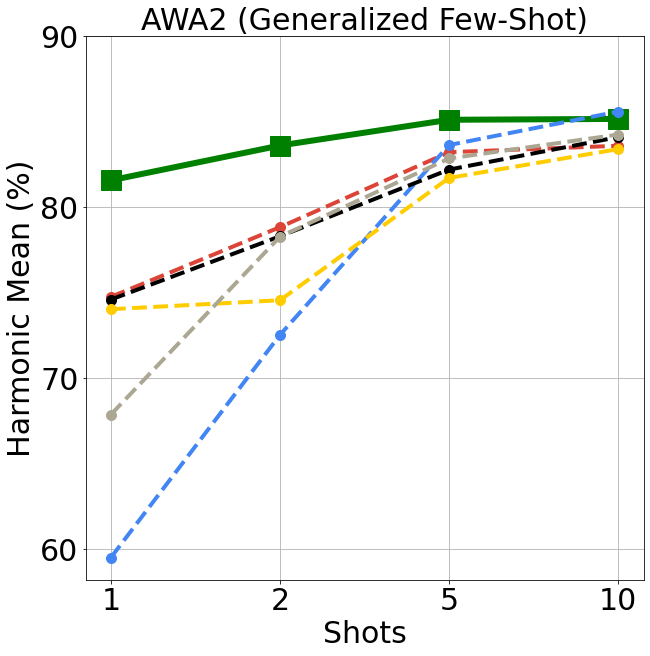}
    \vspace{-5mm}
\vspace{4mm}
\end{minipage}
\begin{minipage}[b]{0.24\linewidth}
  \includegraphics[width=1\textwidth]{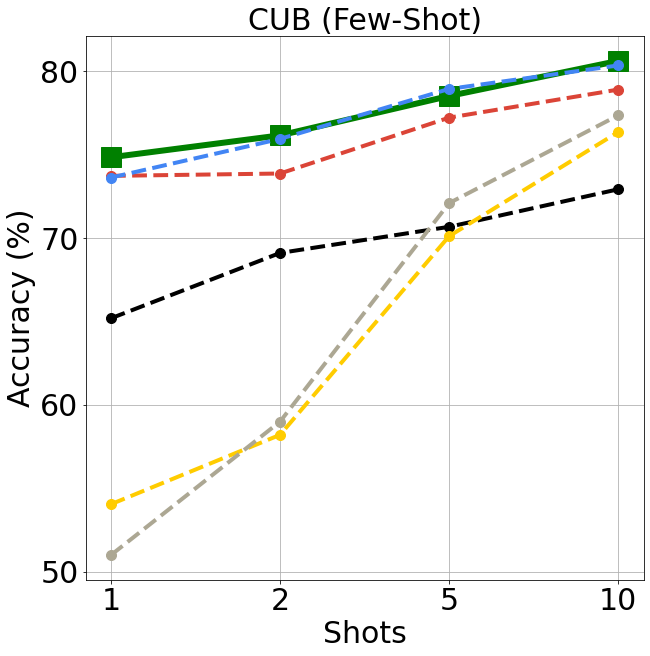}
    \vspace{-5mm}
\vspace{4mm}
\end{minipage}
\begin{minipage}[b]{0.24\linewidth}
  \includegraphics[width=1\textwidth]{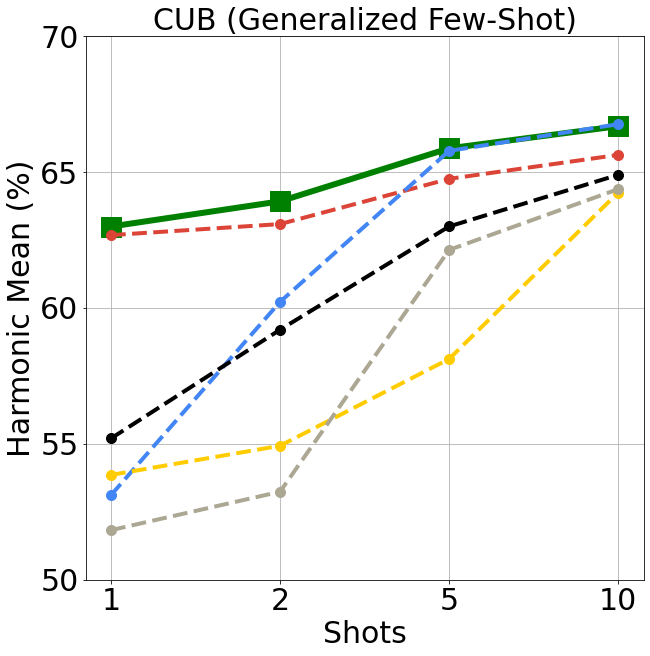}
    \vspace{-5mm}
\vspace{4mm}
\end{minipage}
\vspace{-3.5mm}
\caption{
\small{
Few-shot accuracy ($n\rightarrow n$) and harmonic mean ($H$) for few-shot and generalized few-shot learning in pre-trained setting.
}
\vspace{-5mm}
}
\label{fig:few_shot_performance}
\end{figure*}

\myparagraph{Generalized Zero-Shot Learning:}
Table \ref{tab:gzsl} also shows generalized zero-shot performances. 
We observe that methods using dense features, \texttt{Self-Calibration} and ours, surpass other methods on the majority of datasets.
Specifically, we improve the harmonic mean by at least 5.0\%, 1.7\%, 1.8\% on DeepFashion, AWA2, CUB, respectively, in pre-trained setting and at least 4.6\%, 2.0\% on DeepFashion, AWA2, respectively, in fine-tuned setting.
Our method achieves high accuracy of seen classes and significantly improves accuracy of novel classes by 7.3\%,1.8\%, 6.1\% in pre-trained setting on DeepFashion, AWA2, CUB, respectively.
Notice that fine-tuning on CUB, SUN with small number of samples overfits to training data due to the high capacity of dense attention.

\begin{figure}[h]
\centering
\includegraphics[width=0.80\linewidth]{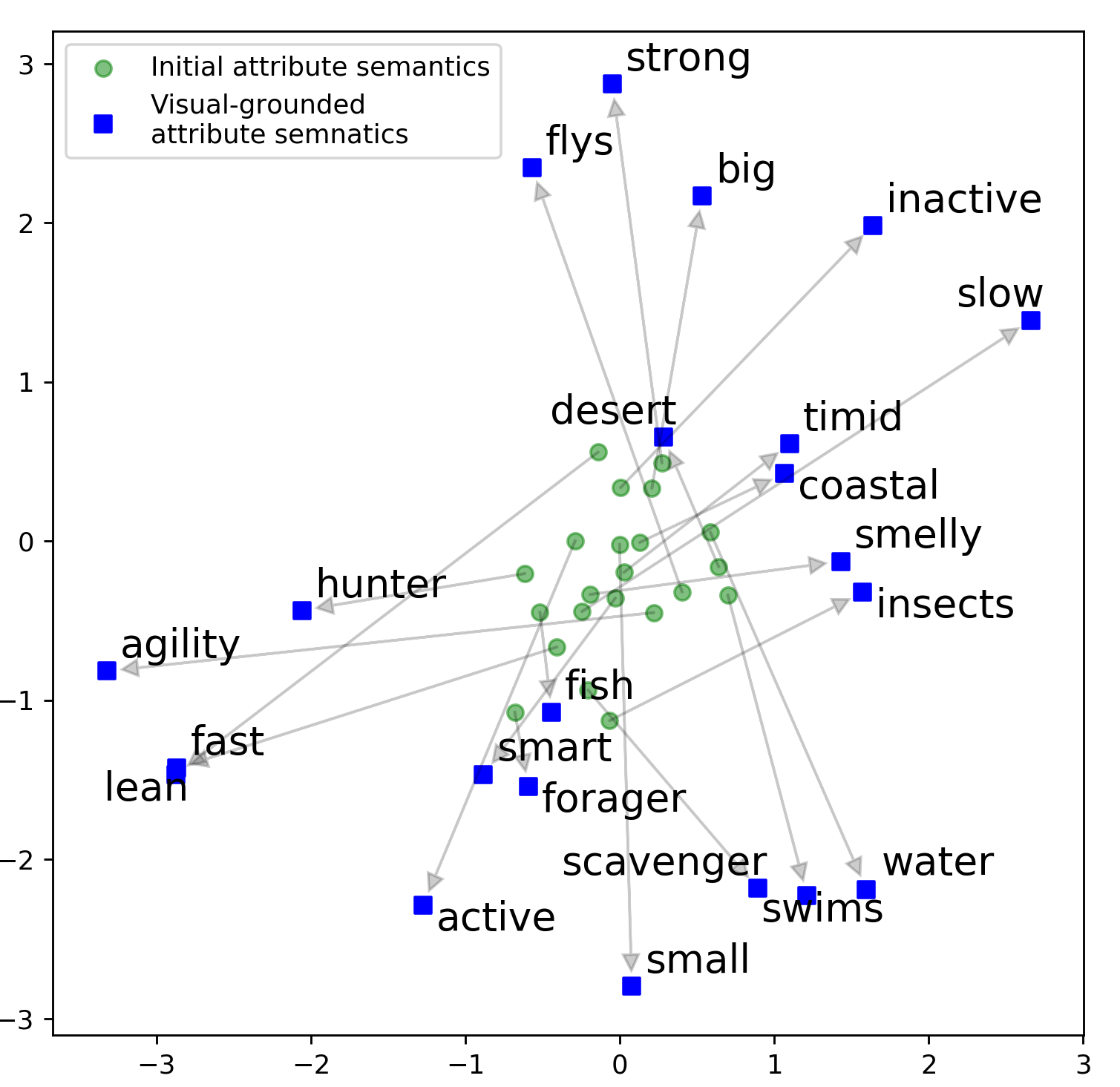}
\vspace{-0mm}
\caption{
\small{
Visualization of a subset of attribute semantic vectors before and after visual grounding on AWA2 dataset.
}
}
\label{fig:visual_grounding}
\end{figure}

\myparagraph{Benefits of Dense Attention Mechanism:}
Table \ref{tab:dense_att_ablation} shows the performances of different attention mechanism variants on generalized zero-shot learning without feature composition.
We observe that without dense attention and attribute grounding, the model performs poorly at both seen and novel class recognition on AWA2 and CUB since it cannot capture fine-grained attribute details.
Although having dense attention without attribute grounding improves CUB performances, this degrades AWA2 performances dues to the misalignment between attribute semantics from word embedding and visual features, as shown in Figure \ref{fig:visual_grounding}.
As such, learning both dense attention and grounding attribute semantics are crucial for achieving high performances. 

\begin{table}
\centering
\resizebox{\linewidth}{!}{

\begin{tabular}{|c|c|ccc|ccc|}
\hline
\multirow{2}{*}{\makecell{Dense\\Attention}}  & \multirow{2}{*}{\makecell{Attribute\\Grounding}} &\multicolumn{3}{c|}{\textbf{AWA2}}&\multicolumn{3}{c|}{\textbf{CUB}}\\
 \cline{3- 8}
 &  & $a\rightarrow s$ & $a\rightarrow n$ & $H$ & $a\rightarrow s$ & $a\rightarrow n$ & $H$ \\
\hline
\hline
No & No & 59.2 & 19.1 & 28.9 & 16.7 & 12.1 & 14.1   \\
Yes  & No & 54.0 & 18.4 & 27.5 & 36.5 & 27.1 & 31.1 \\
Yes  & Yes & \textbf{82.5} &	 \textbf{25.7} & \textbf{39.4} & \textbf{65.3} & \textbf{41.9} & \textbf{51.1}\\
\hline
\end{tabular}
}
\caption{\small{Ablation study of dense attention mechanism without feature composition on the CUB and AWA2 datasets.}}
\label{tab:dense_att_ablation}
\end{table}

\myparagraph{Benefits of Dense Feature Composition:}
We compare recent generative methods and \texttt{Attribute GANs}, where we learn a separate GANs per attribute, with random composition which uniformly samples combinations from $\U(S)$
and our methods in Table \ref{tab:composition_strategies_DeepFashion_ablation} (left).
Random composition significantly outperforms recent methods by at least 1.1\% and 3.4\% harmonic mean on AWA2 and CUB, respectively while performing comparably with \texttt{Attribute GANs}.
We believe this is due to the strong regularization effect of dense feature composition which prevents overfitting on feature combinations from training samples while retaining the ability to recognize fine-grained details obtained from seen classes.
Notice that \texttt{Attribute GANs} cannot scale to 312 attributes in CUB dataset due to its large memory consumption for training hundreds of GANs, thus we do not report its performance.
Our method surpasses random composition by 3.7\% (6.0\%), 2.9\% (2.1\%) in harmonic mean (zero-shot accuracy) on AWA2 and CUB, respectively.

\myparagraph{Effect of Hyperparameters:}
Figure \ref{fig:DeepFashion_hyper_params} shows the zero-shot and generalized zero-shot performances on DeepFashion in the fine-tuned setting as functions of $b$, $\beta$ and $k$. We vary the value of one hyper-parameter while fixing the remaining hyper-parameters and measure the improvement with respect to the lowest value in each hyperparmeter range.
By increasing the search budget for most probable combination via the number of candidate combinations $b$, we improve both zero-shot and generalized zero-shot performances compared to random composition ($b=0$). 
The performances stabilize across a wide range of $b$ as probable samples according to $p(\H|\z)$ often have high $p(y|\H,\z)p(\H|\z)$ probability. Increasing $\beta$ increases the similarity between composed features and features of related samples, thus our method constructs ``hard'' features being closer to the decision boundary of seen classes. The harmonic mean improves with $\beta$ and degrades for large values of $\beta$, as the composed features become too similar to training features.
When increasing the size of the related samples via $k$, we improve zero-shot accuracy the most, as composed features have richer attribute details by using more samples. 
Notice that our method only uses at most $k=20$ related samples, as additional samples are not selected since they do not improve the reconstruction loss \eqref{eq:Q}, thus the performances remain the same for larger $k$.

\myparagraph{Ablation Study:}
We report the effectiveness of different components in our method on DeepFashion in Table \ref{tab:composition_strategies_DeepFashion_ablation} (right). We observe that the discriminative model, trained only on seen classes, fails to generalize to novel classes. 
Although training on random composed features improves the harmonic mean, this does not significantly improve zero-shot accuracy due to the lack of meaningful knowledge in composed features. Using only prior $p(\H|\z)$ improves the harmonic mean by 2.0\% but not zero-shot accuracy, as maximizing the prior is equivalent to using features from the most related sample without modifications for novel classes.
We achieve the best performance when combining the classification knowledge $p(y|\H,\z)$ and the prior knowledge $p(\H|\z)$.
Notice without varying the sample set $S$, the harmonic mean drops by 3.3\%, showing the importance of feature diversity for zero-shot learning.

\begin{figure*}[h]
\centering
\includegraphics[width=0.98\linewidth]{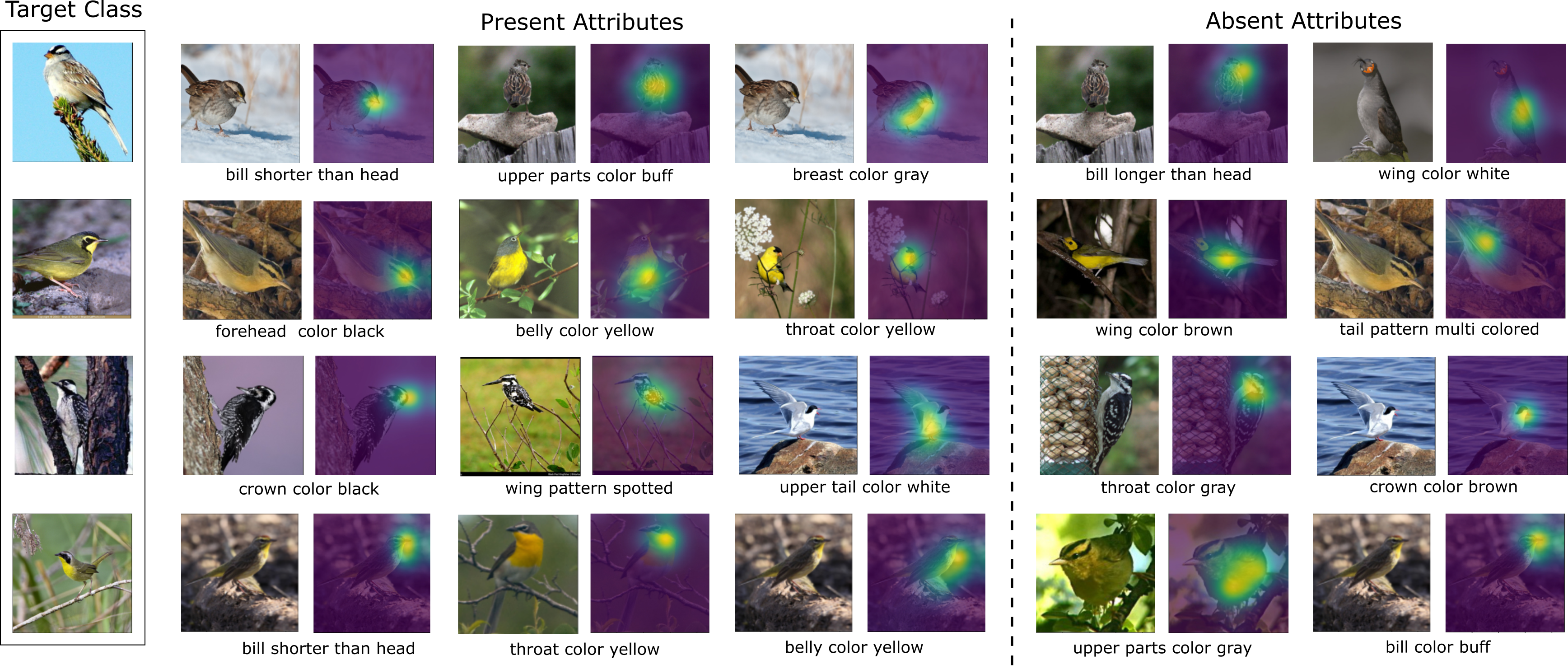}
\vspace{-0mm}
\caption{
\small{Attention visualization of attribute features from samples used for dense feature composition of target classes. Our method selects relevant attributes from related samples to describe novel classes. 
}
}
\label{fig:dense_composition_samples}
\end{figure*}

\myparagraph{Attribute grounding:}
We visualize how the attribute semantic vectors changes during training.
By modifying the word embedding according to \eqref{eq:loss_att}, we observe the semantic representations of related attributes, such as ``lean'', ``fast'' and ``agility'', are brought together. On the other hand, opposite attributes, such as ``fast'' and ``slow'', are separated apart in the semantic space.
Thus, attribute grounding improves performances via aligning visual and textual domains.

\begin{figure}[t!]
\centering
\includegraphics[width=0.72\linewidth]{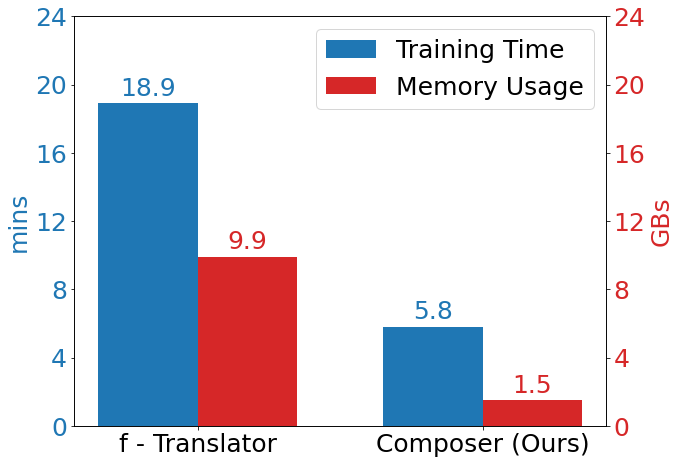}
\vspace{-0mm}
\caption{
\small{
Comparison between \texttt{f-Translator} and our method in terms of time and memory complexities on DeepFashion.
}
}
\label{fig:complexity}
\end{figure}

\myparagraph{Complexity Analysis:}
Figure \ref{fig:complexity} shows memory/space complexity of \texttt{f-Translator} which use a generative model to train a discriminative model, and our method which directly trains a discriminative model. We measure only training time which excludes loading and evaluation times for 6000 iterations and GPU memory usage. By using a single model for feature generation and classification, we improve the training time by 3 folds and the memory usage by 6 folds compared to \texttt{f-Translator}.

\myparagraph{Qualitative Results:}
Figure \ref{fig:dense_composition_samples} visualizes attention maps of attribute features for both present and absent attributes in composed features for novel classes.
In addition to using relevant features for present attributes, our method also selects appropriate features to indicate the absence of attributes in novel classes.
Thus, the discriminative model, trained on these features, learns from both present and absent attributes to recognize novel classes.

\subsection{Few-Shot Experiments}
For fair comparisons, we follow the setting of \cite{Schonfeld:CVPR19}, whose implementation is publicly available.
To be specific, we measure the mean few-shot and generalized few-shot performances over ten runs with one, two, five and ten shots for each novel class on CUB, AWA2 datasets in pre-trained setting.
We compare with three variants of our model as \texttt{Visual-Semantic Composer}, which synthesizes class features based on visual-semantic features $\bar{\h}$, \texttt{Semantic Composer}, which generates features conditioned on only $\z$, and \texttt{No Composer} which trains a discriminative model on only real dense features.

\myparagraph{Few-Shot Learning:}
Figure \ref{fig:few_shot_performance} shows the few-shot performances ($n\rightarrow n$) on both AWA2 and CUB datasets.
We significantly improve the performances by 9.4\% and 3.7\% compared to \texttt{AM3} on AWA2 as well as 9.6\% and 7.0\% compared to \texttt{CADA} in CUB for one- and two-shot settings, respectively.
All variants of our methods significantly outperform \texttt{CADA} due to our ability to capture fine-grained features.
Our \texttt{Visual-Semantic Composer} is most effective on AWA2 dataset where each class is described by a small number of 85 attributes and additional visual information is beneficial for feature generation. 
Thus, we achieve 5.6\%, 5.2\% improvements compared to \texttt{Semantic Composer} in one- and two-shot learning, respectively.
On CUB, as \texttt{Semantic Composer} can leverage a large number of carefully annotated attributes for composition, it can achieve comparable performances to \texttt{Visual-Semantic Composer}.
When given one or two training samples, \texttt{WeightImprint, AM3} have low performances due to overfitting.

\myparagraph{Generalized Few-Shot Learning:}
Figure \ref{fig:few_shot_performance} also shows our method improves the harmonic mean by 7.0\% and 5.3\% on AWA2 and 7.8\% and 4.7\% on CUB compared to \texttt{CADA} given only one and two training samples of novel classes.
Without learning from synthesized features of novel classes, \texttt{No Composer} misclassifies novel class samples as seen classes, thus cannot perform well in generalized few-shot setting despite its high few-shot performances.
This shows the importance of generating novel class features to mitigate the imbalance between seen and novel training samples, which causes seen class bias.

\myparagraph{Effects of Training on Real and Synthesized Features:}
We further measure the effectiveness of training on real and synthesized features by varying $\lambda$ in \eqref{eq:self-compose_fs}.
We observe low few-shot performances when mostly learning from synthesized features ($\lambda=2$) since these samples cannot accurately describe novel classes.
On the other hand, learning from only real features ($\lambda=0$) results in low generalized few-shot performances as the model overfits on few training samples.
This shows the necessity of learning from both real and synthesized features, which prevents overfitting on few training examples and improves generalization toward novel classes.
Our method is most effective when we put more weight on real features compared to synthesize features ($\lambda =0.5$) to train the discriminative model.

\begin{figure}[t!]
\centering
\begin{minipage}[b]{0.45\linewidth}
  \includegraphics[width=1\textwidth]{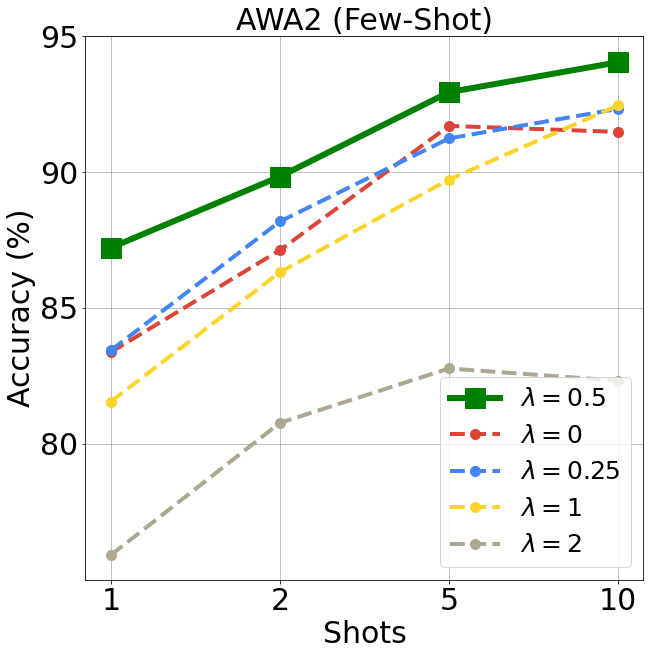}
  \vspace{-1mm}
\end{minipage}
\begin{minipage}[b]{0.45\linewidth}
  \includegraphics[width=1\textwidth]{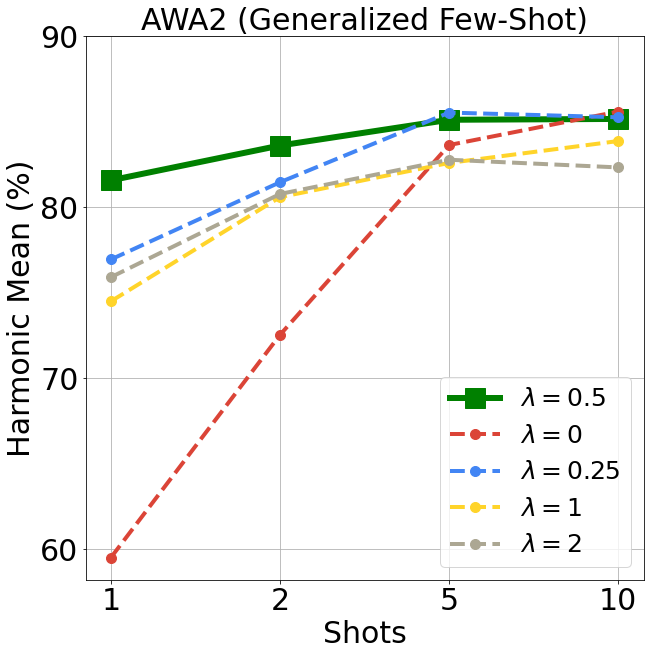}
    \vspace{-1mm}
\end{minipage}
\caption{Effect of training on real and synthesized features for few-shot and generalized few-shot learning.}
\label{fig:few_shot_performance}
\end{figure}

\section{Conclusions}
\label{sec:conclusions}
We proposed a dense feature composition framework that extracts attribute features from training samples and recombines them to construct features of novel classes. Our framework selectively composes features of novel classes from only related training samples and alternates between different samples used for composition to improve the diversity among composed features. We employ a training scheme where a discriminative model composes features to train itself. By extensive experiments on four popular datasets, we show the effectiveness of our method on both zero-shot and few-shot settings.

{\small
\bibliographystyle{IEEEtran}
\bibliography{biblio_bank/ehsan,biblio_bank/multilabellearning,biblio_bank/zeroshot_learning,biblio_bank/fewshot_learning,biblio_bank/recognition,biblio_bank/vision,biblio_bank/sparse,biblio_bank/learning,biblio_bank/neuralnet,biblio_bank/nlp,biblio_bank/generative_model,biblio_bank/finegrained_recognition,biblio_bank/composition,biblio_bank/societal_impact}
}

\begin{IEEEbiography}[{\includegraphics[width=1in,height=1.25in,clip,keepaspectratio]{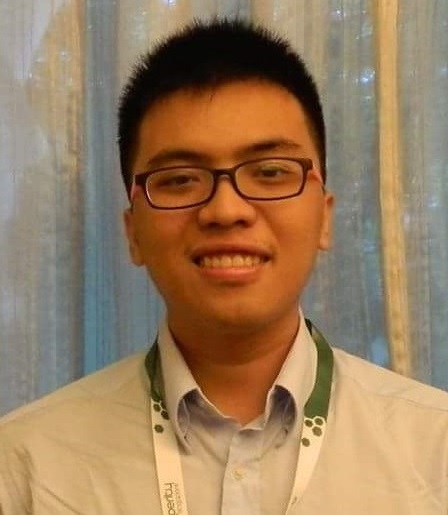}}]%
{Dat Huynh} is a Ph.D. candidate in the Khoury College of Computer Sciences at Northeastern University, advised by Prof. Ehsan Elhamifar. He received his Bachelor's degree from University of Sciences (Viet Nam), where he studied the Advanced Program in Computer Science.
His research interests lie in significantly reducing the amount of training labels for visual classification, detection and segmentation tasks. Specifically, he designs methods that decompose complex concepts into primitive components that can be combined to enable learning with few/zero training samples, with missing annotations and with weak supervision.
\end{IEEEbiography}

\begin{IEEEbiography}[{\includegraphics[width=1in,height=1.25in,clip,keepaspectratio]{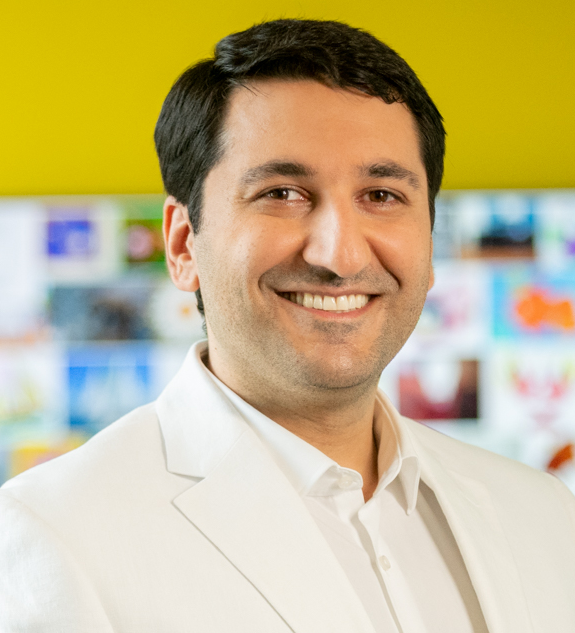}}]%
{Ehsan Elhamifar} is an Assistant Professor in the Khoury College of Computer Sciences and is the director of the Mathematical Data Science (MCADS) Lab at the Northeastern University. Prof. Elhamifar is a recipient of the DARPA Young Faculty Award and the NSF CISE Career Research Initiation Initiative Award. Previously, he was a postdoctoral scholar in the Electrical Engineering and Computer Science (EECS) department at the University of California, Berkeley. Prof. Elhamifar obtained his PhD from the Electrical and Computer Engineering (ECE) department at the Johns Hopkins University. He obtained two Masters degrees, one in Electrical Engineering from Sharif University of Technology in Iran and another in Applied Mathematics and Statistics from the Johns Hopkins University. Prof. Elhamifar’s research areas are computer vision, machine learning and optimization. He develops scalable, robust and interpretable method to address challenges of complex and massive high-dimensional data and applies them to solve real-world challenging problems, including structured data summarization, procedure learning from instructional videos, large-scale and fine-grained recognition with small/no labeled data.
\end{IEEEbiography}

\end{document}